\documentclass[10pt,journal,compsoc]{IEEEtran}

\usepackage{cite}
\usepackage{times}
\usepackage{epsfig}
\usepackage{graphicx}
\usepackage{amsmath}
\usepackage{amssymb}
\usepackage{multirow}
\usepackage{amsthm}
\usepackage{subfigure}
\usepackage{caption}
\usepackage{color}
\usepackage{csquotes}
\usepackage{soul}
\usepackage{enumerate}
\usepackage[normalem]{ulem}
\usepackage{algorithm}
\usepackage{algorithmicx}
\usepackage{algpseudocode}
\usepackage{booktabs}
\usepackage{array}
\usepackage{rotating}
\usepackage{arydshln}
\usepackage{ragged2e}
\usepackage{wrapfig}
\usepackage{hyperref}
\usepackage{amssymb}% http://ctan.org/pkg/amssymb
\usepackage{pifont}% http://ctan.org/pkg/pifont
\usepackage{tabularx}
\usepackage{diagbox}
\usepackage[dvipsnames]{xcolor}

\usepackage[normalem]{ulem}
\newcommand{\xmark}{\ding{55}}
\hypersetup{breaklinks=true,colorlinks}

\newcommand\ie{\emph{i.e.}} 
 
\newcommand\etc{\emph{etc.}}
 
\newcommand\etal{\emph{et al.}}

\definecolor{darkgreen}{RGB}{0,127,0}
\definecolor{darkred}{RGB}{200,0,0}
\def\greencheckmark{\textcolor{darkgreen}{\checkmark}}
\def\redxmark{\textcolor{darkred}{\xmark}}      % pifont
\begin{document}
% \title{COME15K: RGB-D Saliency Detection via Cascaded Mutual Information Minimization}
% \title{Learning Generative Vision Transformer with Energy-Based Prior for Saliency Prediction}
% \title{Generative Vision Transformer for Saliency Prediction with Energy-Based Prior}
% \title{An Energy-Based Prior for Saliency Detection}
% \title{An Energy-Based Prior for Generative Saliency}
\title{Vicinity Vision Transformer}

% \author{Jing~Zhang,~\IEEEmembership{Student Member,~IEEE,}
%       Deng-Ping~Fan*,~\IEEEmembership{Member,~IEEE,}
%       Yuchao~Dai,~\IEEEmembership{Member,~IEEE,}
%       Saeed~Anwar,~\IEEEmembership{Member,~IEEE,}
%       Fatemeh Sadat~Saleh,~\IEEEmembership{Member,~IEEE,}
%       Sadegh~Aliakbarian
%       and
%       Nick~Barnes,~\IEEEmembership{Member,~IEEE,}%
      
\author{Weixuan Sun,
       Zhen Qin,
       Hui Deng,
       Jianyuan Wang,
       Yi Zhang,
       Kaihao Zhang,\\
       Nick Barnes,
       Stan Birchfield,
       Lingpeng Kong,
       Yiran Zhong
 \IEEEcompsocitemizethanks{
% \IEEEcompsocthanksitem Weixuan Sun is with School of Computing, the Australian National University, Canberra, Australia and Sensetime research, Shanghai, China. 
\IEEEcompsocthanksitem Weixuan Sun, Kaihao Zhang, Nick Barnes are with School of Computing, the Australian National University, Canberra, Australia.
\IEEEcompsocthanksitem Zhen Qin, and Yi Zhang are with SenseTime research, Shanghai, China. 
\IEEEcompsocthanksitem Hui Deng is with the School of Electronics and Information, Northwestern Polytechnical University, Xi'an, China.
\IEEEcompsocthanksitem Jianyuan Wang is with Visual Geometry Group, University of Oxford, Oxford, United Kingdom.
\IEEEcompsocthanksitem Stan Birchfield is with Nvidia, Redmond, WA, USA. 
\IEEEcompsocthanksitem Lingpeng Kong is with the University of Hong Kong, Hong Kong, China.
\IEEEcompsocthanksitem Yiran Zhong is with Shanghai AI Lab, Shanghai, China. 
\IEEEcompsocthanksitem Weixuan Sun and Zhen Qin are with equal contributions.
\IEEEcompsocthanksitem 
Our code is available at: \url{https://github.com/OpenNLPLab/Vicinity-Vision-Transformer}.
\IEEEcompsocthanksitem Corresponding author: Yiran Zhong (zhongyiran@gmail.com).
}
}
\markboth{Journal of \LaTeX\ Class Files,~Vol.~14, No.~8, August~2015}%
{Shell \MakeLowercase{\textit{et al.}}: Bare Demo of IEEEtran.cls for Computer Society Journals}





\IEEEtitleabstractindextext{%
\begin{abstract}
\justifying
Vision transformers have shown great success on numerous computer vision tasks. 
However, their central component,  softmax attention, prohibits vision transformers from scaling up to high-resolution images, due to both the computational complexity and memory footprint being quadratic. 
Linear attention was introduced in natural language processing (NLP) which reorders the  self-attention mechanism to mitigate a similar issue, but directly applying existing linear attention to vision may not lead to satisfactory results.
We investigate this problem and point out that existing linear attention methods ignore an inductive bias in vision tasks, \ie, 2D locality.
In this paper, we propose Vicinity Attention, 
which is a type of linear attention that integrates 2D locality.
Specifically, for each image patch, we adjust its attention weight based on its 2D Manhattan distance from its neighbouring patches. 
In this case, 
we achieve 2D locality in a linear complexity 
where the neighbouring image patches receive stronger attention than far away patches. 
In addition, we propose a novel Vicinity Attention Block that is comprised of Feature Reduction Attention (FRA) and Feature Preserving Connection (FPC) in order to address the computational bottleneck of linear attention approaches, including our Vicinity Attention, whose complexity grows quadratically with respect to the feature dimension.
The Vicinity Attention Block computes attention in a compressed feature space with an extra skip connection to retrieve the original feature distribution. 
We experimentally validate that the block further reduces computation without degenerating the accuracy. 
Finally, to validate the proposed methods, we build a linear vision transformer backbone named Vicinity Vision Transformer (VVT).
Targeting general vision tasks,
we build VVT in a pyramid structure with progressively reduced sequence length.
We perform extensive experiments on CIFAR-100, ImageNet-1k, and ADE20K datasets to validate the effectiveness of our method. 
Our method has a slower growth rate in terms of computational overhead than previous transformer-based and convolution-based networks when the input resolution increases.
In particular, our approach achieves state-of-the-art image classification accuracy with $50\%$ fewer parameters than previous approaches. 
\end{abstract}

}

\maketitle

\IEEEdisplaynontitleabstractindextext

%%%%%%%%% BODY TEXT
\IEEEpeerreviewmaketitle

\IEEEraisesectionheading{\section{Introduction}}
\label{intro_sec}
\IEEEPARstart{R}{ecent}
years have witnessed the success of the transformer structure in natural language processing~\cite{vaswani2017attention,brown2020language,devlin2018bert} and computer vision~\cite{dosovitskiy2020image,He_2022_CVPR,dpt_transformer}.
However, transformers inherently suffer from quadratic computational complexity and a quadratic memory footprint.
As a result, vision transformer networks have to adopt patch-wise image tokenization 
to reduce the sequence length. 
Despite the temporary relief provided by such a tokenization method, the quadratic complexity problem still exists. 
This limitation prohibits vision transformers from handling high-resolution images or fine-grained image patches.

Linear attention is a promising direction to solve this issue. 
This group of methods reorders the self-attention mechanism of transformers with Kernelization methods
to reduce the quadratic complexity to linear~\cite{tay2022efficient}.
Numerous methods have been proposed for attention decomposition, such as approximating the softmax~\cite{xiong2021nystromformer,peng2021random,choromanski2020rethinking}, and finding a new similarity metric~\cite{katharopoulos2020transformers,zhen2022cosformer}.
However, most of these methods are only verified on NLP tasks and suffer from a crucial performance drop in computer vision \cite{lu2021soft}, when compared with conventional softmax attention. 

We investigate this issue and point out that existing linear attention methods
ignore an inductive bias in vision, \ie, 2D locality. 
Our intuition is based on the fact that convolutional neural networks~\cite{simonyan2014very,he2016deep_resnet,krizhevsky2012imagenet} have dominated computer vision tasks since the rise of deep networks. 
Most of them have such locality bias, that is, 2D neighbouring regions are more likely to be highly related.
Recent transformer-based methods also adopt such an assumption and 
obtain improved results by 
attaching convolution-based 2D locality bias \cite{wu2021cvt,han2021transformer,yuan2021tokenstotoken,dai2021coatnet,srinivas2021bottleneck}.
Additionally, some efficient transformer backbones use window attention~\cite{liu2021swin,vaswani2021scaling}, neighbourhood attention~\cite{zhang2021multiscale}, or deformable attention~\cite{zhu_deformableDETR_ICLR_2021,xia2022vision} to enable lower complexity and 2D locality, but they still suffer limited receptive field and quadratic complexity within the sampled tokens.
Therefore we hypothesize that the 2D locality bias is an important property and should be incorporated into linear transformers.

\begin{figure*}[t]
% \centering
   \begin{center}
   {\includegraphics[width=0.8\linewidth]{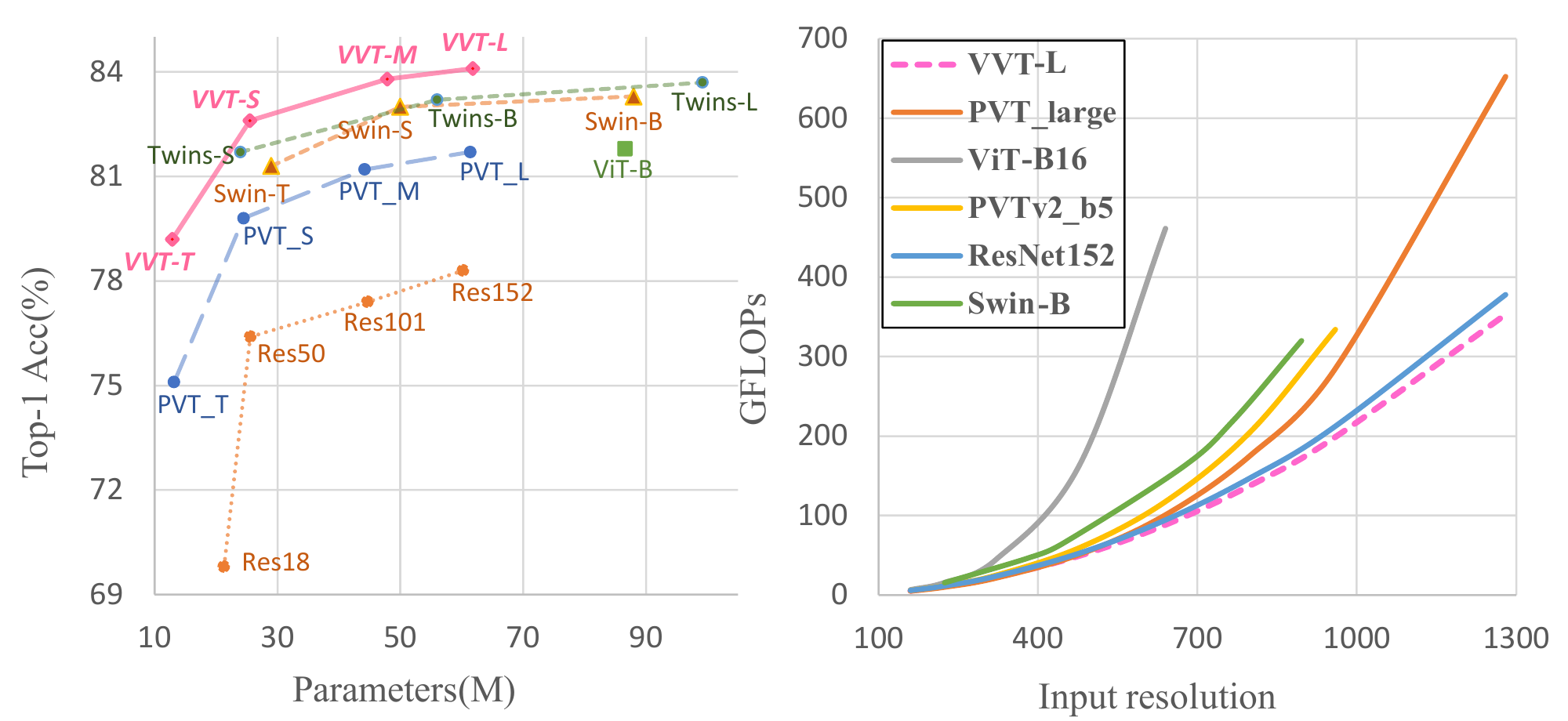}} 
   \end{center}
   % \vspace{-5mm}
\caption{Top-1 accuracy of our Vicinity Vision Transformer (VVT) with respect to parameters on the ImageNet-1k\cite{krizhevsky2012imagenet} dataset, and the GFLOPs corresponding to various input image sizes. Our VVT outperforms all competitors with $50\%$ fewer parameters, and it enjoys the lowest GFLOPs growth rate.
}
\vspace{-3mm}
\label{fig:intro}
\end{figure*}

In this paper,
we present Vicinity Attention, a new linear attention method that effectively enforces 2D locality.
Our locality re-weighting mechanism is inspired by the recently proposed cosFormer~\cite{zhen2022cosformer}, 
which assumes locality bias in language and uses a cosine re-weighting mechanism for 1D NLP tasks. 
However, directly applying the cosFormer to vision tasks will lead to unsatisfactory results since the 1D locality enforces stronger connections only on the 1D tokenized neighbouring image patches, so the cosFormer is not compatible with 2D distance. 
In that case, it will assign less weight to vertically connected patches as they are far away when we tokenized the patches to 1D tokens as shown in Fig.~\ref{fig:reweight}a. 
To solve this issue, we propose a 2D variant of cosFormer in this work. 
Specifically, 
we propose a 2D Manhattan distance decomposition to encode relative positions, and integrate it with the cosine function
to encourage visual tokens to have a stronger connection to their neighbors in 2D (Fig.~\ref{fig:reweight}b-d). 
Since this distancing mechanism is decomposable in two directions, it can be seamlessly applied to linear attention.

Compared with vanilla transformer-based methods, linear attention complexity grows linearly with respect to sequence length but quadratically with feature dimension, which becomes a new computational bottleneck in our Vicinity Attention. 
To address this concern so as to further reduce computation, we propose a novel Vicinity Attention Block.
First, we propose Feature Reduction Attention (FRA) to reduce the input feature dimension by half, so that the overall theoretical complexity can be reduced by a factor of four. 
Then a Feature Preserving Connection (FPC) is added to retrieve the original feature distribution and strengthen the representational ability. 
Such a block structure is seamlessly integrated with our linear Vicinity Attention and 
we experimentally validate that our Vicinity Vision Block further reduces computational complexity without degenerating the accuracy.

Finally, we build a general-purpose linear vision backbone, termed Vicinity Vision Transformer (VVT).
Since our linear Vicinity Attention has a better efficiency advantage over 
vanilla self-attention\cite{dosovitskiy2020image},
which enables feature maps with higher resolutions.
We build VVT in a pyramid structure, which starts from high-resolution image patches and progressively shrinks to adapt to different vision tasks with multi-scale outputs. 

Fig.~\ref{fig:intro} a compares our method with current state-of-the-art vision backbones on the ImageNet-1k benchmark. 
Our method outperforms all the competitors with only half the parameters. 
We also provide the growth rate of GFLOPs with different input resolutions for these methods in Fig.~\ref{fig:intro} b. 
Given its linear complexity in token numbers, VVT can efficiently process images with much larger resolutions.
Further, our experiments validate that VVT does not compromise accuracy and achieves superior results over transformer-based methods as well as convolution-based methods of comparable model sizes.

Our main contributions are as follows:
(1) We introduce a linear self-attention mechanism for vision called Vicinity Attention, which introduces a Manhattan distance-based 2D locality to linear vision transformers. 
(2) To further reduce computational complexity by targeting linear attention,  
we propose a novel attention block called the Vicinity Attention Block.  
It contains a feature reduction attention~(FRA) to improve the efficiency and a feature preserving connection~(FPC) to retain the feature extraction ability. 
(3) We correspondingly build the Vicinity Vision Transformer~(VVT), which serves as a general-purpose vision backbone and can be easily applied to various vision tasks. 
Extensive experiments validate the effectiveness of VVT on various computer vision benchmarks.

\begin{figure*}[t]
   \begin{center}
   {\includegraphics[width=0.8\linewidth]{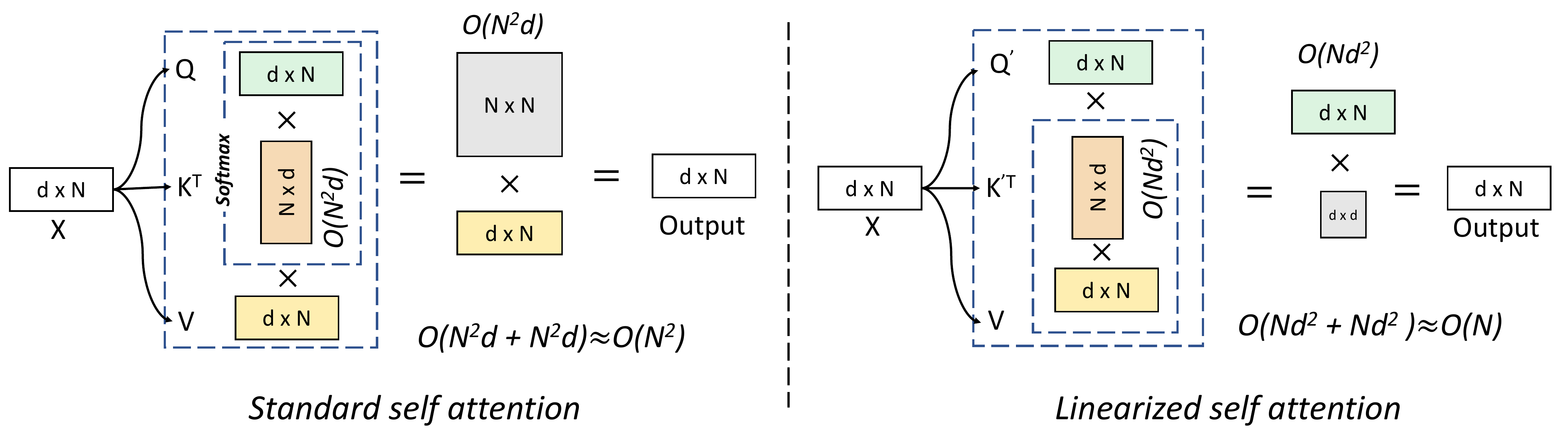}} 
   \end{center}
   \vspace{-5mm}
\caption{Illustration of the standard self-attention (left) and linearized self-attention (right). 
$N$ denotes the patch number of the input image, and $d$ is the feature dimension. 
With $N\gg d$, the computational complexity of the linearized self-attention grows linearly with respect to the input length, while that of the standard self-attention is quadratic.}
\label{fig:linear}
\end{figure*}

\section{Related Work}
\subsection{Vision Transformer Backbones}
In computer vision tasks,
CNN networks\cite{simonyan2014very,he2016deep_resnet,krizhevsky2012imagenet} have achieved great successes, while recently transformers have gained strong emerging interest. In this section, we mainly discuss vision networks using self-attention.

Some works~\cite{hu2019local,ramachandran2019stand,zhao2020exploring,srinivas2021bottleneck} adopt self-attention mechanisms to replace some or all convolution layers in the CNN networks for image recognition.
To further leverage the power of self-attention, ~\cite{Wang_2018_CVPR} proposes a non-local operation and adds it within ResNet~\cite{he2016deep}, the non-local block can capture long-range dependencies and lead to improvements on vision tasks such as video classification, object detection,
segmentation, and pose estimation.
Recently, pure transformer-based vision networks were introduced.
ViT~\cite{dosovitskiy2020image} splits images into local patches. Then it projects and flattens patches into embedding sequences, subsequently, it employs a pure transformer structure for image classification. 

Variations of ViT like CVT~\cite{wu2021cvt}, Swin~\cite{liu2021swin}, PVT~\cite{Wang_2021_ICCV}, Twins~\cite{chu2021twins} and T2T~\cite{yuan2021tokenstotoken} were proposed.
CVT~\cite{wu2021cvt} introduces Convolutional Token Embedding and Convolutional Projection for Attention to include  desirable properties of CNNs into 
ViT~\cite{dosovitskiy2020image} to improve both performance and efficiency.
The Swin Transformer~\cite{liu2021swin} introduces non-overlapping windows and applies self-attention in each window, then the window partitions are shifted between adjacent layers. 
Swin transformer reduces computational complexity and is suitable for different vision tasks.
PVT~\cite{Wang_2021_ICCV} introduces a pure transformer network with a pyramid structure, and it further proposes spatial reduction attention which computes self-attention in reduced sequence length to save computation.
Twins~\cite{chu2021twins} combines locally-grouped self-attention from Swin~\cite{liu2021swin} and sub-sampled attention from \cite{Wang_2021_ICCV}, which decreases the computational cost and enhances communications between sub-windows.
DAT~\cite{xia2022vision} builds a general vision transformer backbone with deformable attention.
More recently, \cite{tang2022quadtree} adopts the quad-tree algorithm which progressively ignores less related image patches to improve efficiency. 
However, none of the above methods are able to directly process full-length self-attention on high-resolution inputs,
requiring locally grouping or sub-sampling. 
Contrarily, our method can directly calculate self-attention on high-resolution input.

\subsection{Efficient Transformers}
Various works have been proposed to address the computational complexity problems of transformers in both computer vision and natural language processing. In addition to vision backbones such as CVT~\cite{wu2021cvt}, Swin~\cite{liu2021swin}, PVT~\cite{Wang_2021_ICCV}, and Twins~\cite{chu2021twins} introduced in the previous section, in this section we introduce other more efficient transformer algorithms.

Existing efficient transformer methods can be generally grouped into two categories:  pattern-based and kernel-based.
Pattern-based methods sparsify the attention matrix with handcrafted or learnable patterns.
~\cite{child2019generating} reduces the complexity by applying a combination of a strided pattern and a local pattern to the standard attention matrix. 
Beyond fixed diagonal windows and global windows, 
Longformer ~\cite{beltagy2020longformer} also extends sliding windows with dilation to enlarge the receptive field. 
Instead of fixed patterns, Reformer~\cite{kitaev2020reformer} and ~\cite{daras2020smyrf} adopt locally sensitive hashing to group tokens into different buckets.
On the other hand, kernel-based methods ~\cite{peng2021random,choromanski2020rethinking,wang2020linformer,xiong2021nystromformer} aim to replace softmax self-attention with approximations or decomposable functions, which change the order of scale dot product calculation and reduce the complexity of self-attention from quadratic to linear.
~\cite{zhen2022cosformer} proposes a cosine re-weighting function to enforce locality in a 1D sequence and achieves linear complexity. Nevertheless, the aforementioned kernel-based methods only consider the linearization in a 1D sequence for natural language processing tasks, their performances on 2D vision tasks are not satisfactory. In contrast, we aim at enforcing 2D locality in linear complexity.

Two works that are similar to ours are \cite{lu2021soft} and \cite{shen2021efficient}.
\cite{shen2021efficient} applies softmax on $Q$ and $K$ respectively, then changes the dot product order to access a linear complexity. However, it is not validated on a large-scale classification benchmark as a backbone network.
SOFT~\cite{lu2021soft} uses a Gaussian kernel function to replace the dot-product similarity in self attention. It assumes that the similarity is symmetrical, which may not hold true in practice. 
Further, it does not consider the 2D locality mechanism.
In this paper, we propose a linear self-attention that facilitates a 2D locality mechanism, and we validate the proposed method on various vision tasks.

\begin{figure*}[t]
   \begin{center}
   {\includegraphics[width=0.8\linewidth]{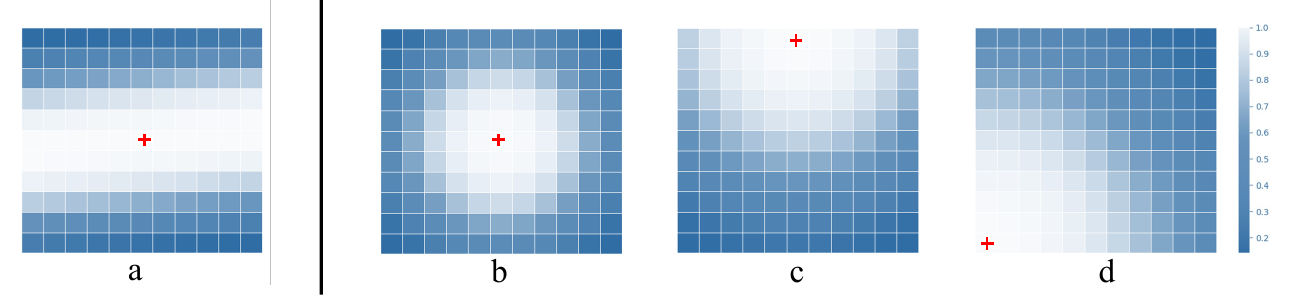}} 
   \end{center}
\caption{Visualization of the locality re-weighting patterns. The red cross symbol denotes the query position and the mask denotes the locality weights to it.
We use a lighter color to indicate a larger weight.
\textbf{(a)} If we directly adopt 1D distance re-weighting \cite{zhen2022cosformer} in self-attention, the query token will be assigned lower weights on vertical tokens as they are far away when processed in 1D.
\textbf{(b), (c), (d)} In our method, 
although self-attention is calculated in 1D, our re-weighting mechanism ensures that a token is encouraged to have higher relation weights with its neighbours in two dimensions. }
\label{fig:reweight}
\end{figure*}

\section{Preliminary}
\textbf{Self-attention} was originally proposed for 1D NLP tasks~\cite{vaswani2017attention}. 
To make it suitable for 2D vision tasks, we often convert an image to several patches and then embed them to a 1D sequence $x \in \mathbb{R}^{N \times d}$, where $d$ is the embedding size and $N = \frac{H}{p} \times \frac{W}{p}$ is the patch number, with $p$, $H$, $W$ denoting patch size, image height, and image width respectively. 
Mathematically, a transformer block $\mathcal{T}:\mathbb{R}^{N\times d}\to \mathbb{R}^{N \times d}$ with input $x$ is defined as:
\begin{equation}
\textstyle{
    \mathcal{T}(x) = \mathcal{F}(\mathsf{Att}(x)+x)
}\end{equation} 
\noindent where $\mathsf{Att}(\cdot)$ is the self-attention module and $\mathcal{F}(\cdot)$ is a feed-forward module.

The self-attention module $\mathsf{Att}(\cdot)$ adopts three learnable weights $W_Q, W_K, W_V$ to project $x$ into \emph{query} ($Q$), \emph{key} ($K$), and \emph{value} ($V$).
It usually computes an attention matrix $A$ by a similarity function $\mathcal{S}(\cdot)$ over queries and keys. 
In standard self-attention, $\mathcal{S(\cdot)}$ is softmax normalization. %, and hence we have
The output of the self-attention module is  $\mathcal{O}=\mathsf{Att}(x)=AV$, where $A \in \mathbb{R}^{N \times N}$ suffers from a quadratic space and time complexity with respect to the patch number $N$.
Therefore, the theoretical computation complexity here is $\textstyle O(N^{2}d)$ = $O(\frac{H^{2}W^{2}}{p^{4}}d)$.
Consequently the self-attention module becomes sensitive to image size $H,W$ and patch size $p$, which are the computation bottlenecks in vision transformer networks.

\noindent\textbf{Linearization} of  self-attention aims to reduce the quadratic theoretical computation complexity to linear. 
It can be achieved by picking a decomposable similarity function $\mathcal{S}(\cdot)$ to satisfy
\begin{equation}
    \mathcal{S}(Q_i, K_j) = \phi(Q_i)\phi(K_j)^\top
    \label{eq: decomp}
\end{equation}
where $\phi(\cdot)$ is a kernel function.
Given such a kernel, we can write the outputs of the self-attention module as follows:
\begin{equation}
   \mathcal{O}_i = \sum_j \frac{\mathcal{S}(Q_i,K_j)}{\sum_j \mathcal{S}(Q_i,K_j)}V_j = \frac{\sum^{N}_{j=1}(\phi (Q_i) \phi({K_j})^\top) V_j}{\sum^{N}_{j=1}(\phi (Q_i) \phi({K_j})^\top)}
    \label{eq: generalized att}
\end{equation}

Fig. \ref{fig:linear} illustrates the linearization process. Instead of calculating the attention matrix $A \in \mathbb{R}^{N \times N} $, we compute $\phi({K})^\top V \in\mathbb{R}^{d \times d}$ first:
\begin{equation}
\textstyle
        [\phi (Q) \phi({K})^\top] \ V = \phi (Q) \ [\phi({K})^\top V]
    \label{eq: linear form}
\end{equation}
\noindent so that the $O(N^{2}d)$ operation is converted to an $O(Nd^{2})$ one.
Its computational complexity grows linearly with respect to the sequence length $N$. 
A number of linear attention approaches have been proposed for NLP tasks, which use different kernel functions $\phi(x)$ to replace the quadratic softmax, as discussed in the Related Work. 
However, 
directly applying existing linear attention\cite{zhen2022cosformer,choromanski2020rethinking,xiong2021nystromformer} to vision suffers from a performance drop\cite{lu2021soft}.
Notably, all of the aforementioned linear attention approaches and existing linear vision transformers\cite{lu2021soft,tang2022quadtree,ali2021xcit}
do not consider 2D locality in vision.
To address this problem, we propose a linear self-attention that is aware of 2D position. 

\noindent\textbf{Locality} is a widely used assumption in computer vision~\cite{ Wang_2021_CVPR, cheng2019noise, wu2021cvt, yuan2021tokenstotoken, dai2021coatnet, han2021transformer, he2016deep_resnet, simonyan2014very, krizhevsky2012imagenet}, \emph{i.e.,} neighbouring pixels should have a higher possibility to belong to the same object than distant pixels. 
In convolution-based networks, this assumption is inherently coupled into each layer throughout the whole model via 
convolutional kernels~\cite{krizhevsky2012imagenet, simonyan2014very, he2016deep_resnet}. 
However, it does not hold for standard transformer-based networks due to the self-attention mechanism~\cite{dosovitskiy2020image}. 
Several works show that with the same number of parameters, transformer-based networks cannot match the performance of the CNN counterparts, possibly due to the lack of locality bias~\cite{wu2021cvt,chu2021we}.
Recent state-of-the-art methods partially introduce locality bias to vision transformers at the architectural level.
\cite{wu2021cvt,han2021transformer,yuan2021tokenstotoken,dai2021coatnet} directly combine convolution with transformers. 
Some efficient transformer backbones use window attention~\cite{liu2021swin,vaswani2021scaling}, neighbourhood attention~\cite{zhang2021multiscale}, or deformable attention~\cite{zhu_deformableDETR_ICLR_2021,xia2022vision} to achieve lower complexity and better performance.
These methods enable 2D locality, 
however, they still suffer limited receptive field and quadratic complexity within the sampled tokens.

In this paper, for the first time, we introduce locality bias into the linear self-attention.
It can be smoothly integrated into existing vision transformer architectures for better performance. 
In fact, our method
achieves better performance than previous state-of-the-art vision transformers in various computer vision benchmarks. 

\section{Our Method}
\begin{figure*}[htb]
   \begin{center}
   {\includegraphics[width=0.8\linewidth]{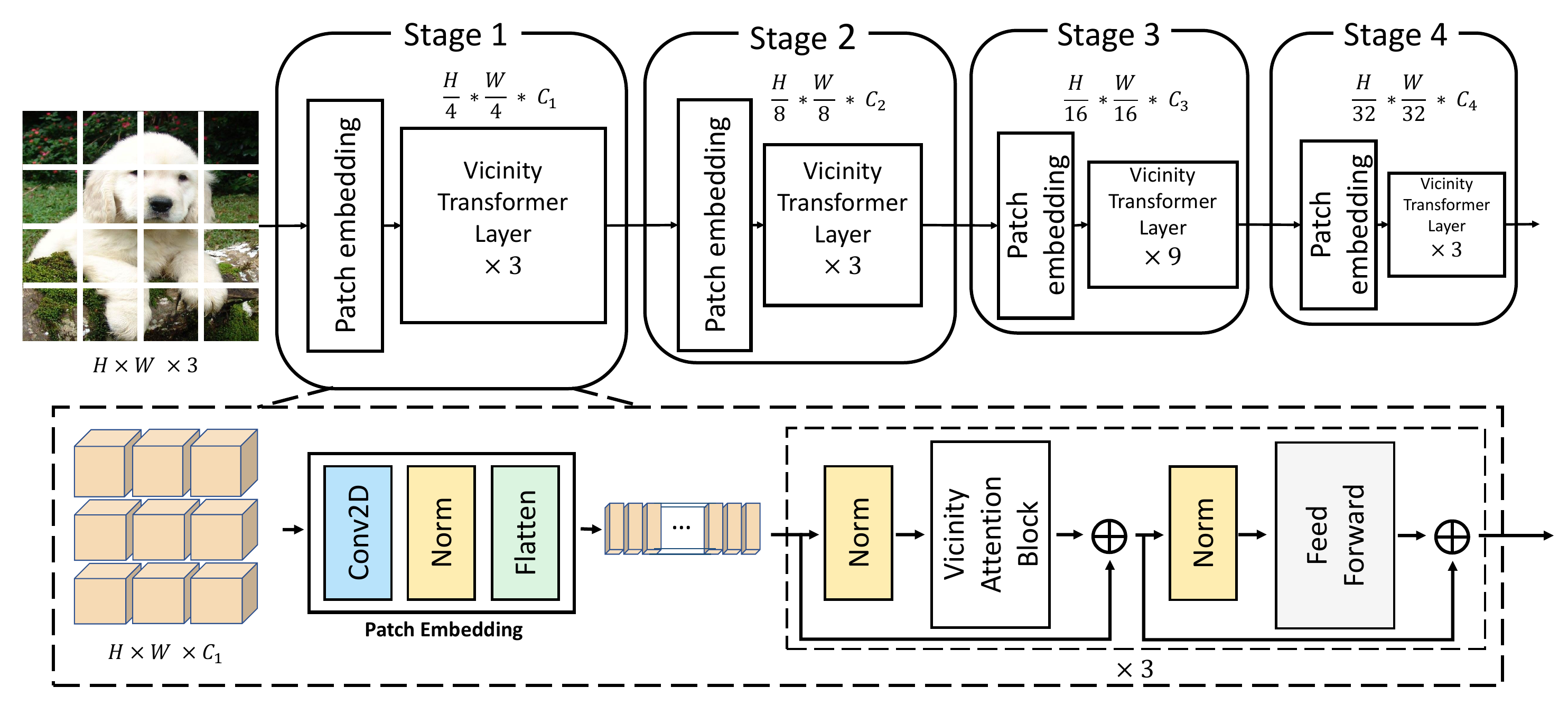}} 
   \end{center}
   \vspace{-5mm}
\caption{Overall architecture of our VVT. VVT adopts a pyramid structure and is divided into four stages. The illustrated sample is VVT-Small. The output resolution  progressively shrinks to generate multi-scale feature maps.}
\label{fig:overview}
\end{figure*}

In this section, we start from presenting the details of Vicinity Attention mechanism in Sec.~\ref{sec:vi_att} which enforces 2D locality in linear attention.
Then in Sec.~\ref{sec:attention block}, we introduce a new attention structure named Vicinity Attention Block to address the computational bottleneck for linear attention 
targeting linear attention.
In Sec.~\ref{sec:vvt structure} we introduce the overall architecture of VVT and show the structure illustration in Fig.~\ref{fig:overview}.
Targeting general vision tasks, we integrate our Vicinity Attention into a four-stage pyramid structure. 
Each stage consists of a patch embedding module followed by several Vicinity Vision Blocks and feed forward residual blocks, 
Image inputs are progressively down-sampled to generate multi-scale outputs. 
We provide a family of VVT variants and detail their configurations in Table.~\ref{table: variants}.

\subsection{Vicinity Attention}
\label{sec:vi_att}
It has been verified that the softmax normalization is the root of the quadratic complexity of self-attention~\cite{lu2021soft}.
In this section, we provide the details of our linear self-attention technique, \ie, vicinity attention for vision tasks.
The key insight is to replace the standard softmax operation by another similarity function, which (1) can be decomposed to simple kernel functions and (2) introduces a locality constraint that is beneficial in visual understanding.
Inspired by \cite{zhen2022cosformer}, we adopt $\mathrm{ReLU}(\cdot)$ as the kernel function, and conduct a row-wise normalization to replace softmax.  
Our self-attention then can be reformulated as:

\begin{equation}
%\textstyle{
\label{eq: relu attention}
\begin{aligned}
    \mathcal{O}_i = \frac{\sum^{N}_{j=1}[\mathrm{ReLU} (Q_i) \mathrm{ReLU}({K_j})^\top] V_j}{\sum^{N}_{j=1}[\mathrm{ReLU} (Q_i) \mathrm{ReLU}({K_j})^\top]} \\
    = \frac{\mathrm{ReLU} (Q_i) \sum^{N}_{j=1}[\mathrm{ReLU}({K_j})^\top V_j]}{\mathrm{ReLU} (Q_i)\sum^{N}_{j=1} \mathrm{ReLU}({K_j})^\top}
\end{aligned}
\end{equation}
where we can reorder the calculation so the complexity is reduced to linear with respect to the sequence length $N$. 
This $\mathrm{ReLU}$-based normalization method retains two important properties of softmax self-attention: (1) it is always positive, avoiding
aggregating negatively-correlated information. (2) all the elements lie between $[0,1]$.

\noindent\textbf{Enforcing 2D Locality Bias}
Locality bias has been discussed in language tasks \cite{zhen2022cosformer,clark2019does,kovaleva2019revealing} with 1D sequence distance encoding, but not considered by existing linear attention in vision. 
In vision transformers, the embedding sequence is flattened from a 2D mask, hence it is essential to consider token positions in 2D before being flattened.
To enforce the locality bias in linear transformers, we need a kernel function that can 1) put more emphasis on 2D neighboring patches and 2) can be decomposed with Eq.~\eqref{eq: decomp}.

Given two tokens $q_i$, $k_j$ from $Q$ and $K$ respectively, the positions of these two tokens on the 2D feature maps before flattening are:
\begin{equation}
    i=u_{i}m + r_i, \ j=u_{j}m + r_j, \quad 0 \leq r_i, r_j < m
    \label{eq: position}
\end{equation}
where $m$ is width of the feature map, $u$ denotes the row index,  and $r$ denotes the column index. 
Following Eq.~\eqref{eq: decomp}, we can define a re-weighted attention with a distance function $\mathcal{D}$ to enforce the locality bias between two tokens as:
\begin{equation}
    {\mathcal{S}(Q_i,K_j)} = \phi (Q_i)\mathcal{G}(\mathcal{D}(i,j))\phi({K_j})^\top
\end{equation}

\noindent where $\mathcal{G}$ produces the weight according to the distance so nearby patches can be emphasized.
A naive choice might be directly using the Euclidean distance $\sqrt{(r_{j}-r_i)^{2}+(u_j-u_i)^{2}}$.
However, since this term cannot be decomposed into two terms relating to $i$ and $j$ separately, it cannot be applied to the linear transformers.

Instead, since 2D Manhattan distance~\cite{minkowski1910geometrie} decouples relative position in two directions, we propose to use it as the distance function:

\begin{equation}
    \label{equation: manhattan distance}
    {\mathcal{S}(Q_i,K_j)}   = \phi (Q_i)\mathcal{G}(|r_j-r_i|+|u_j-u_i|)\phi({K_j})^\top
\end{equation}
\noindent
However, direct Manhattan distance decoupling is still hindered as the absolute operation cannot be decomposed. 

Inspired by~\cite{zhen2022cosformer}, the cosine function has two desirable features: 1) It cancels the absolute operation in Eq.~\eqref{equation: manhattan distance} and applies non-linear emphasis on the nearby tokens; and 2) It can be decomposed into two terms relating to $i$ and $j$ separately, which fulfills the linear complexity.
Thus, we propose to bind Manhattan distance and cosine function to achieve linear attention with 2D locality.
First, given a feature map with size $m$ by $n$, following Eq.~\eqref{eq: position} we redefine the 2D positions as:
\begin{equation}
    a_i = \frac{\pi{u_i}}{2n}, \ a_j = \frac{\pi{u_j}}{2n},  \ b_i = \frac{\pi{r_i}}{2m},  \ b_j = \frac{\pi{r_i}}{2m}
\end{equation}
\noindent Then, using the Manhattan distance, the self-attention calculation can be decomposed into four terms:
\begin{equation}
\label{eq: locality_decomp}
\begin{aligned}
    {\mathcal{S}(Q_i,K_j)} 
    =\, &\phi (Q_i)(\cos(a_i-a_j)+\cos(b_i-b_j))\phi({K_j})^\top\\
    =\,  &\phi (Q_i)(\cos(a_i)\cos(a_j))\phi({K_j})^\top 
    + \\ &\phi (Q_i)(\sin(a_i)\sin(a_j))\phi({K_j})^\top +\\
    & \phi (Q_i)(\cos(b_i)\cos(b_j))\phi({K_j})^\top + \\&\phi (Q_i)(\sin(b_i)\sin(b_j))\phi({K_j})^\top \\
    =\,  & \phi^{\cos a_i}(Q_i)\phi^{\cos a_j}(K_j)^\top + \\&\phi^{\sin a_i}(Q_i)\phi^{\sin a_j}(K_j)^\top + \\
     & \phi^{\cos b_i}(Q_i)\phi^{\cos b_j}(K_j)^\top + \\&\phi^{\sin b_i}(Q_i)\phi^{\sin b_j}(K_j)^\top 
    \end{aligned}
\end{equation}
where $\textstyle
    \phi^{\cos y}(x) \triangleq \cos(y) \phi(x), \phi^{\sin y}(x) \triangleq \sin(y) \phi(x) $.
    
Here the queries and keys are related to their own positions $i$ and $j$, and hence we can fulfill Eq.~\eqref{eq: linear form} to reorder the dot product in self-attention.
Following our locality decomposition, we get:
\begin{equation}
\begin{aligned}
Q_i'&= [\phi^{\cos a_i}(Q_i), \phi^{\sin a_i}(Q_i), \phi^{\cos b_i}(Q_i), \phi^{\sin b_i}(Q_i)] \\
K_j'&= [\phi^{\cos a_j}(K_j), \phi^{\sin a_j}(K_j), \phi^{\cos b_j}(K_j),\phi^{\sin b_j}(K_j)] 
\end{aligned}
\end{equation}
where
\begin{equation}
\begin{aligned}
\phi^{\cos y}(x) \triangleq \cos(y) \phi(x), \phi^{\sin y}(x) \triangleq \sin(y) \phi(x)
\end{aligned}
\end{equation}

Finally, we have the output as:
\begin{equation}
\begin{aligned}
    \mathcal{O}_i 
    &= \frac{\sum^{N}_{j=1}[ Q_i' {K_j'}^\top] V_j}{\sum^{N}_{j=1}Q_i' {K_j'}^\top}
    &=\frac{ Q_i' \sum^{N}_{j=1}[{K_j'}^\top V_j]}{Q_i'\sum^{N}_{j=1} {K_j'}^\top}\\
\end{aligned}
\end{equation}

Our method provides three appealing properties: 
(1) \textit{Linear complexity}: cosine re-weighting and Manhattan distance ensure decomposability, thus we avoid calculating $QK$. 
The overall computational complexity is $O(N(4d)^2)$ which grows linearly with sequence length.
(2) \textit{Locality bias}: if two tokens are close to each other in 2D feature maps, they are encouraged to have a stronger relationship, as shown in Fig.~\ref{fig:reweight}.
(3) \textit{Global context}: it 
retains a global receptive field as standard self-attention.

\begin{figure*}[htb]
   \begin{center}
   {\includegraphics[width=0.7\linewidth]{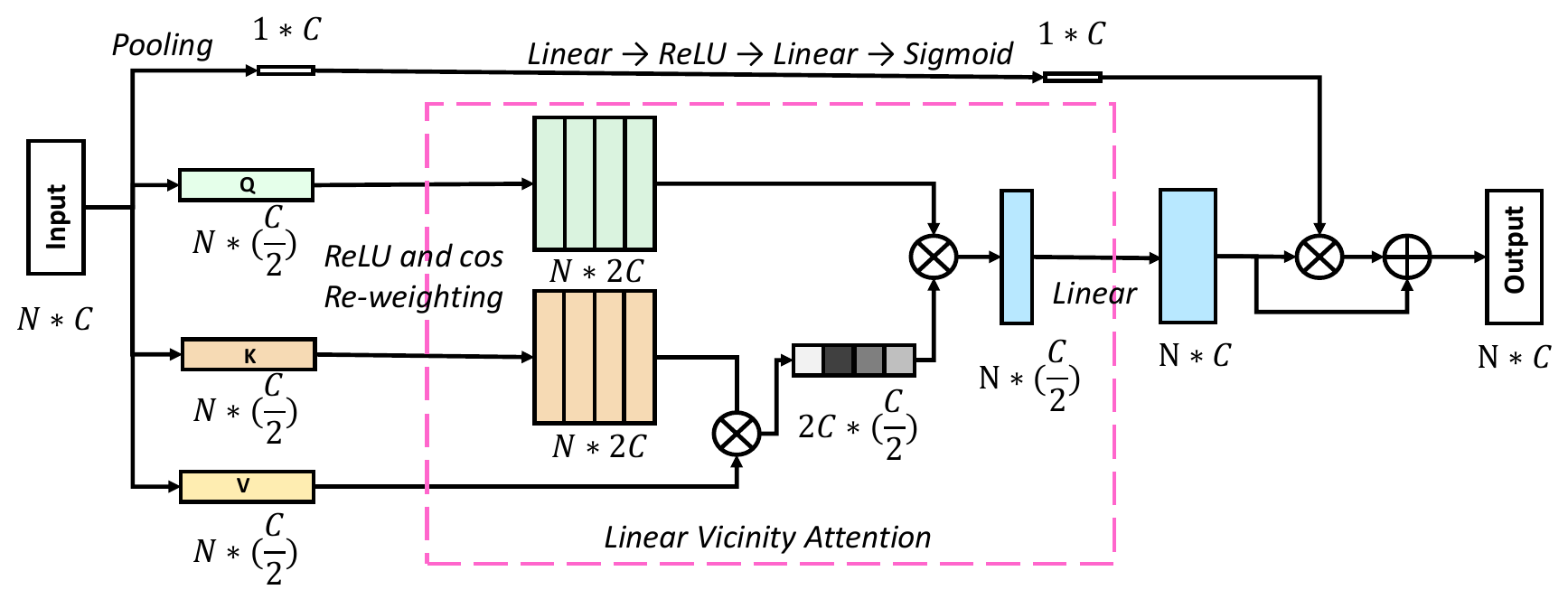}} 
   \end{center}
   \vspace{-5mm}
\caption{Illustration of the Vicinity Attention Block. 
The input feature dimension is down-sampled to reduce the computational cost, while the original feature distribution is retained by the FPC (the skip connection at the top).}
\label{fig:transformer block}
\end{figure*}

\begin{table*}[htb]
% \footnotesize
\centering  
\caption{\textbf{Detailed specifications of our VVT variants.} We show four variants of our VVT structure. C: Feature dimension, P: Patch size, R: Feature reduction, H: Head number, E: Expansion ratio of the feed-forward layer. 
The number of layers of the four stages follows the 
ratios of $1:1:3:1$ or $1:1:9:1$.}
\centering\scalebox{1}{
\begin{tabular}{c|c|c|c|c|c|c}
                         & Output Size       & Layer Name                                                     & Tiny                                                          & Small & Medium & Large \\ \hline
\multirow{2}{*}{Stage 1} & \multirow{2}{*}{$\frac{H}{4} \times \frac{W}{4}$} & Patch Embedding                                                & \multicolumn{4}{c}{C = 96, P = 4}                                                     \\ \cline{3-7} 
                         &                   & \begin{tabular}[c]{@{}l@{}}Transformer \\ Encoder\end{tabular} & \begin{tabular}[c]{@{}l@{}}	$\begin{bmatrix}R=2\\ H = 1\\E = 8 \end{bmatrix} \times 2$ \end{tabular} &   $\begin{bmatrix}R=2\\ H = 1\\E = 8 \end{bmatrix} \times 3$    &     $\begin{bmatrix}R=2\\ H = 1\\E = 8 \end{bmatrix} \times 3$   &   $\begin{bmatrix}R=2\\ H = 1\\E = 4 \end{bmatrix} \times 4$    \\ \hline
\multirow{2}{*}{Stage 2} & \multirow{2}{*}{$\frac{H}{8} \times \frac{W}{8}$} & Patch Embedding                                                & \multicolumn{4}{c}{C = 160, P = 2}                                               \\ \cline{3-7} 
                         &                   & \begin{tabular}[c]{@{}l@{}}Transformer \\ Encoder\end{tabular} &      $\begin{bmatrix}R=2\\ H = 2\\E = 8 \end{bmatrix} \times 2$                                                         &   $\begin{bmatrix}R=2\\ H = 2\\E = 8 \end{bmatrix} \times 3$    &     $\begin{bmatrix}R=2\\ H = 2\\E = 8 \end{bmatrix} \times 3$   &  $\begin{bmatrix}R=2\\ H = 2\\E = 4 \end{bmatrix} \times 4$     \\ \hline
\multirow{2}{*}{Stage 3} & \multirow{2}{*}{$\frac{H}{16} \times \frac{W}{16}$} & Patch Embedding                                                & \multicolumn{4}{c}{C = 320, P = 2}                                                    \\ \cline{3-7} 
                         &                   & \begin{tabular}[c]{@{}l@{}}Transformer \\ Encoder\end{tabular} &     $\begin{bmatrix}R=2\\ H = 5\\E = 4 \end{bmatrix} \times 2$                                                          &    $\begin{bmatrix}R=2\\ H = 5\\E = 4 \end{bmatrix} \times 9$   &   $\begin{bmatrix}R=2\\ H = 5\\E = 4 \end{bmatrix} \times 27$     &   $\begin{bmatrix}R=2\\ H = 5\\E = 4 \end{bmatrix} \times 36$    \\ \hline
\multirow{2}{*}{Stage 4} & \multirow{2}{*}{$\frac{H}{32} \times \frac{W}{32}$} & Patch Embedding                                                & \multicolumn{4}{c}{C = 512, P = 2}                                                    \\ \cline{3-7} 
                         &                   & \begin{tabular}[c]{@{}l@{}}Transformer \\ Encoder\end{tabular} &         $\begin{bmatrix}R=2\\ H = 8\\E = 4 \end{bmatrix} \times 2$                                                      & $\begin{bmatrix}R=2\\ H = 8\\E = 4 \end{bmatrix} \times 3$      &   $\begin{bmatrix}R=2\\ H = 8\\E = 4 \end{bmatrix} \times 3$     &
                         $\begin{bmatrix}R=2\\ H = 8\\E = 4 \end{bmatrix} \times 4$ \\
\hline
\end{tabular}}
\label{table: variants}
\end{table*}

\subsection{Vicinity Attention Block}
\label{sec:attention block}
Compared to vanilla attention, the complexity of linear attention grows linearly with sequence length but quadratically with respect to feature embedding size.
Thus, the computational bottleneck of linear attention is shifted from the input resolution to feature dimension, whilst it is not considered by existing linear vision transformers\cite{lu2021soft,tang2022quadtree,shen2021efficient}.
This issue is amplified in Vicinity Attention as our
theoretical computational complexity is $O(N(4d)^2)$.
To achieve a better computational efficiency in linear attention, 
we redesign the structure of the multi-head self-attention (MSA) module to reduce the feature dimension without hindering the performance.

We present a schematic illustration of our Vicinity Attention Block in Fig.~\ref{fig:transformer block}. Our block is generally composed of two steps: compressing and preserving. 
In the first compressing step,
it receives an input $X \in\mathbb{R}^{N \times d}$, which is projected into a compressed feature space $Q \in\mathbb{R}^{N \times \frac{d}{2}}$, $K \in\mathbb{R}^{N \times \frac{d}{2}}$, $V \in\mathbb{R}^{N \times \frac{d}{2}}$. 
We call it Feature Reduction Attention~(FRA). 
Then, following our locality decomposition in Eq.~\eqref{eq: locality_decomp}, we obtain $Q^{'} \in\mathbb{R}^{N \times 2d}$, $K^{'} \in\mathbb{R}^{N \times 2d}$ which is used to calculate linear self-attention. 
Due to the essence of linear attention, FRA reduces the computational complexity by a factor of four.

In the second preserving step, we propose to preserve the original feature distribution to compensate the feature compression with negligible computational overhead.
Inspired by \cite{Cao_2019_ICCV,hu2018squeeze},
we add a skip connection called Feature Preserving Connection~(FPC) to capture the global context features of input $X$.
The FPC consists of an average pooling operation and two linear layers which have a complexity of $2d^2$.
FPC retains the original feature distribution and can strengthen the representational ability.
Finally, the outputs of FRA and FPC are fused in a residual manner to get final attention output.

In summary, our Vicinity Attention Block proposes to calculate self-attention in a compressed feature dimension while still preserving the original feature space.
It reduces the theoretical complexity to $O(N(2d)^2 + 2d^2)$.
Our experiments validate that it notably reduces computation without degenerating accuracy.

In the following, we discuss the relationships between our VVT and several existing methods.

\noindent\textbf{Relationship to cosFormer.}
The cosFormer~\cite{zhen2022cosformer} was originally
developed for natural language processing and achieves linear complexity in 1D.  In this paper we develop a 2D variant of the cosFormer.
The cosFormer~\cite{zhen2022cosformer} attention adopts cosine re-weighting and considers 1D locality bias in NLP, but it shows poor results on vision tasks (Table \ref{ablation table:attention}).
Our Vicinity Attention can be seen as a non-trivial extension of the cosFormer to vision. 
The combination of Manhattan distance and the cosine function composes a novel 2D positional encoding that enables the capture of long-range visual dependency with 2D locality and linear complexity.
In addition, we propose a new vision attention block named the Vicinity Attention Block, which contains
FRA and FPC to further reduce the complexity without diminishing performance, 
The Vicinity Attention Block is then used to build vision backbones with a pyramid structure and can process high-resolution images.

\noindent\textbf{Relationship to existing window attention and neighborhood attention.}
2D locality has been introduced into vision transformers by several methods such as window attention~\cite{liu2021swin,vaswani2021scaling}, deformable attention~\cite{zhu_deformableDETR_ICLR_2021,xia2022vision} and neighborhood attention~\cite{hassani2022neighborhood}.
A key difference between these and our Vicinity Attention lies in the receptive field. 
These existing methods apply hard locality, \ie, they consider self-attention only within the sampled tokens (from nearby windows or deformable operations) unsampled tokens are disregarded. Further, complexity within the sampled tokens is still quadratic. 
Differently, Vicinity Attention has a constant global receptive field via soft locality, \ie, all spatial locations are visible, but spatially nearby ones are emphasized via our linear attention.

\noindent\textbf{Relationship to GCNet.}
The main difference between the Vicinity Attention Block and GCNet block~\cite{Cao_2019_ICCV} is the attention generation method, which reflects the different motivations of the two blocks.
The GCNet
block adds a uniform context vector to every spatial location. This uniform context vector does not vary by query location and is motivated by the observation that different attention maps are mostly similar.
Our Vicinity Attention Block performs a transformer operation that is modified to reduce complexity. That is, our attention maps vary by query location using the 2D locality constraint.
We also have a channel attention-like connection called FPC which is an instantiation of the squeeze-excitation (SE) connection~\cite{hu2018squeeze}, but aims to reduce complexity while preserving  representation ability. 
Results in Table~\ref{ablation table:attention} show the superiority of the Vicinity Attention Block compared to the GCNet block.

\subsection{The Vicinity Vision Transformer}
\label{sec:vvt structure}
In this section, we introduce the overall architecture of Vicinity Vision Transformer (VVT) as shown in Fig.~\ref{fig:overview}.

The linear complexity of Vicinity Attention enables higher-resolution inputs, so 
we integrate the Vicinity Attention Block into a progressively shrinking pyramid structure that has four stages to generate feature maps at different scales. 
Each stage contains a patch embedding layer and multiple stacked Vicinity Transformer Blocks and Feed-forward blocks.
In detail, we use a patch size of $4 \times 4$ in the first stage. 
Given the input image of size $H \times W \times 3$, we first divide it into $\frac{H}{4} \times \frac{W}{4}$ patches. 
Then we feed the patches into a patch embedding module to obtain a flattened embedding sequence with a size of $\frac{HW}{4^2} \times C_{1}$. 
Here we adopt the overlapping patch embedding module and Convolutional Feed-Forward proposed by Wang \etal~\cite{wang2022pvtv2}.
The embedding sequence is subsequently fed into several successive transformer blocks with $L_1$ layers.

\begin{figure}[htb]
  \begin{center}
    \includegraphics[width=0.4\textwidth]{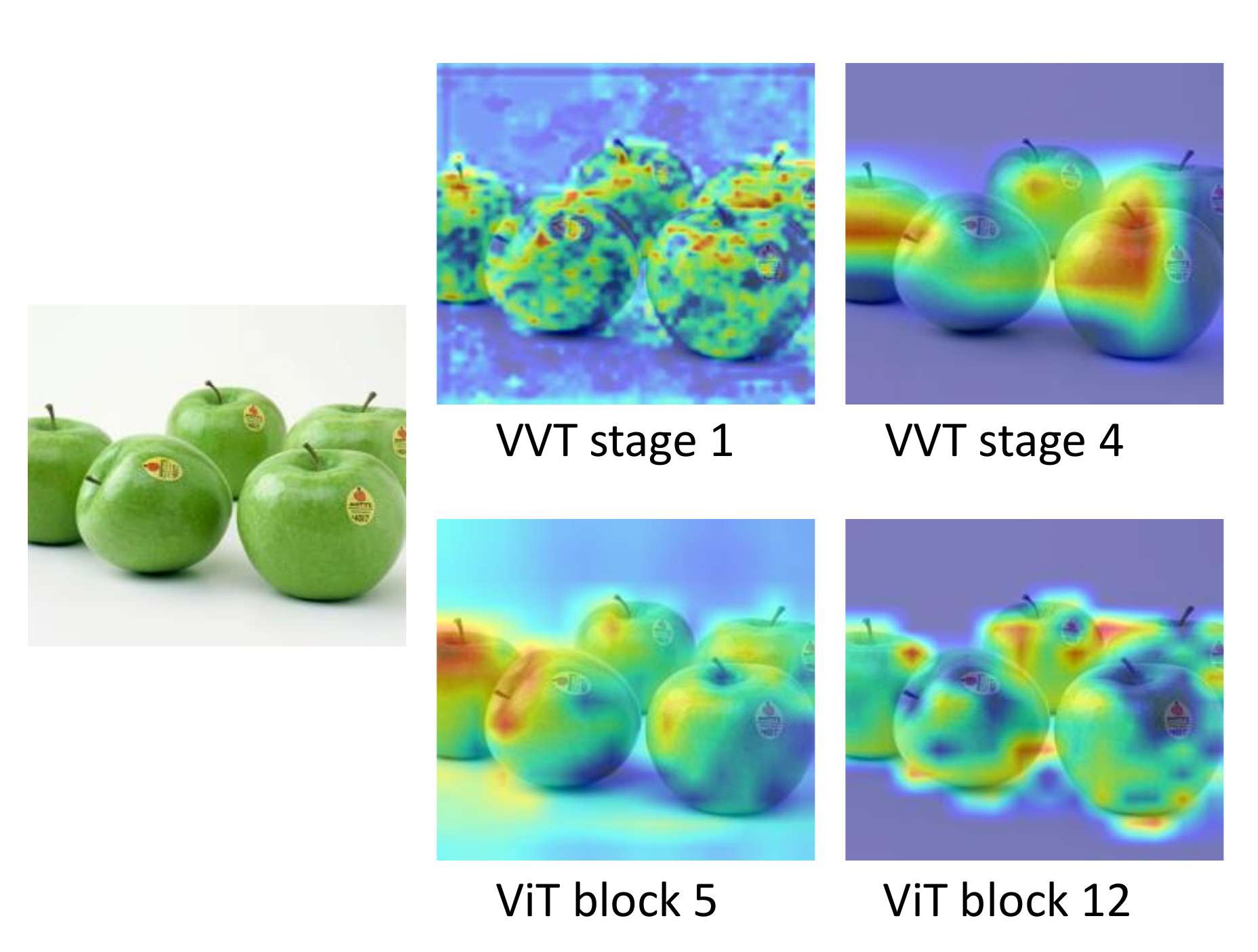}
  \end{center}
  \vspace{-5mm}
  \caption{Grad-CAM of the proposed network and ViT\cite{dosovitskiy2020image}. Our network can generate high-resolution feature maps in early stages, which contain more fine-grained local object features.
  }
  \label{fig:cam}
\end{figure}

In the second stage, the feature sequence from the first stage is reshaped back to $\frac{H}{4} \times \frac{W}{4} \times C_{1}$ and down-sampled to an embedding sequence of size $\frac{HW}{8^2} \times C_{2}$, and then processed by the transformer blocks of the second stage. 
We follow the same approach to obtain multi-scale output feature maps of the third and fourth stage with output sizes of  $\frac{H}{16} \times \frac{W}{16}$ and $\frac{H}{32} \times \frac{W}{32}$, respectively.
Hierarchical feature maps can be easily leveraged to many downstream vision tasks.
In Fig.~\ref{fig:cam}, we show qualitative examples of Grad-CAM \cite{selvaraju2017grad} obtained from different stages of  VVT and ViT \cite{dosovitskiy2020image} respectively.
As shown,
our approach is able to produce fine-grained features whereas the ViT \cite{dosovitskiy2020image} can only capture low-resolution features.

Table \ref{table: variants} details different architecture variants of VVT from VVT\_Tiny to VVT\_Large. 
We empirically choose our network settings 
following the common principles of vision backbones\cite{he2016deep_resnet,wang_PVT_2021}: (1) spatial resolution is decreased progressively with feature dimension increased. (2) stage 3 has the most of the computational cost.
The architecture hyper-parameters of VVT are:
\begin{itemize}
	\item C: the input channel dimension of the attention block.
	\item P: the patch size of the patch embedding.
	\item R: the feature reduction ratio of the Vicinity Attention Block.
	\item H: the number of heads in the Vicinity Attention Block.
	\item E: the feature expansion ratio of the feed forward layer.
\end{itemize}
\vspace{-5mm}

\begin{table*}[h]
%\footnotesize
\tabcolsep=0.7cm
\centering
\caption{\textbf{Comparison of different backbones on ImageNet1k\cite{deng2009imagenet}.} For a fair comparison, we compare VVT with competing models under similar parameter sizes.
All models are trained and tested on $224 \times 224$ resolution. GFLOPs are also calculated under the input scale of $224 \times 224$. ``MS out" represents multi-scale outputs. 
}
%\resizebox{0.9\linewidth}{90mm}{
\begin{tabular}{r|c|c|c|c|c}
% \toprule
Method & Style  & MS out & Param & GFLOPs & Top-1 Acc(\%)  \\
\hline
ResNet18\cite{he2016deep} & ConvNets  & $\greencheckmark$ & 11.7 & 1.8 & 69.8 \\
DeiT-Tiny/16\cite{touvron2021training}& Transformers & \redxmark & \textbf{5.7} & \textbf{1.3} & 72.2 \\
PVTv2-B1\cite{wang2022pvtv2}& Transformers  & $\greencheckmark$ & 13.1 & 2.1 & 78.7 \\
SOFT \cite{lu2021soft}& Transformer  & $\checkmark$ & 13.0 & 1.9 & 79.3 \\
VVT-T(ours)& Transformers   & $\greencheckmark$ & 12.9 & 3.0 & \textbf{79.4}  \\
\hline
ResNet50\cite{he2016deep}& ConvNets  & $\greencheckmark$  &25.6 &  4.1 & 76.1  \\
PVT-Small\cite{Wang_2021_ICCV}& Transformers  & $\greencheckmark$ & 24.5 & 3.8 & 79.8 \\
RegNetY-4G\cite{xu2021regnet}& ConvNets  & $\greencheckmark$ & 21.0 & 4.0  &  80.0 \\
TNT-S\cite{han2021transformer}& Transformers  & \redxmark & 23.8 & 5.2 & 81.3 \\
Swin-T\cite{liu2021swin}& Transformers  & $\greencheckmark$ & 29.0 & 4.5 & 81.6\\
CvT-13\cite{wu2021cvt}& Hybrid  & $\greencheckmark$ & \textbf{20.0} & 4.5 &81.6 \\
Twins-SVT-S\cite{chu2021twins}& Hybrid  & $\greencheckmark$& 24.0& \textbf{2.8} & 81.7 \\
XCiT-S12\cite{ali2021xcit} & Linear Transformers  & \redxmark  & 26.0 & 4.8 & 82.0\\ 
DAT-T\cite{xia2022vision} & Transformer & \greencheckmark & 29.0 & 4.6 & 82.0 \\
PVTv2-B2-Li\cite{wang2022pvtv2}& Transformers  & $\greencheckmark$ & 22.6 & 3.9 & 82.1 \\
SOFT\cite{lu2021soft}& Linear Transformers  & $\greencheckmark$ & 24.0 & 3.3 & 82.2 \\
Focal-Tiny\cite{yang2021focal} & Transformers  & $\greencheckmark$ & 29.1 & 4.9 & 82.2 \\
Quadtree-B-b2 \cite{tang2022quadtree}& Linear Transformers & $\greencheckmark$ & 24.2 & 4.5 & 82.7 \\
VVT-S(ours)& Linear Transformers  & $\greencheckmark$ & 25.5 & 5.6 & \textbf{82.7} \\
\hline
ResNet101\cite{he2016deep}& ConvNets  & $\greencheckmark$ &44.7 &7.9 &77.4 \\
ViT-Small16\cite{dosovitskiy2020image}& Transformers  & \redxmark & 48.8 & 9.9 & 80.8 \\
PVT-Medium\cite{Wang_2021_ICCV}& Transformers  & $\greencheckmark$ & 44.2 & \textbf{6.7} & 81.2 \\
RegNetY-8G\cite{xu2021regnet}& ConvNets  & $\greencheckmark$ & 39.0 & 8.0 & 81.7 \\
CvT-21\cite{wu2021cvt}& Hybrid  & $\greencheckmark$ &\textbf{32.0} & 7.1 & 82.5\\
XCiT-S24\cite{ali2021xcit} & Linear Transformers  & \redxmark & 48.0 & 9.1 & 82.6\\ 
SOFT\cite{lu2021soft}& Linear Transformers  & $\greencheckmark$ & 45.0 & 7.2 & 82.9 \\
Swin-S\cite{liu2021swin}& Transformers  & $\greencheckmark$ & 50.0 & 8.7 & 83.0 \\
Twins-SVT-B\cite{chu2021twins}& Hybrid  & $\greencheckmark$ & 56.0 & 8.3 & 83.2 \\
PVTv2-B3\cite{wang2022pvtv2}& Transformers  & $\greencheckmark$ & 45.2 & 6.9 & 83.2 \\
Focal-Small\cite{yang2021focal} & Transformers & $\greencheckmark$ & 51.1 & 9.1 & 83.5\\
PVTv2-B4\cite{wang2022pvtv2}& Transformers  & $\greencheckmark$ & 62.6 & 10.1 & 83.6 \\
DAT-S\cite{xia2022vision} & Transformer & \greencheckmark & 50.0 & 9.0 & 83.7 \\
Quadtree-B-b3 \cite{tang2022quadtree}& Linear Transformers & $\greencheckmark$ & 46.3 & 7.8 & 83.7 \\
VVT-M(ours) & Linear Transformers  & $\greencheckmark$ & 47.9 & 9.4 & \textbf{83.8} \\ 
\hline
ResNet152\cite{he2016deep}& ConvNets  & $\greencheckmark$ & \textbf{60.2} & 11.6 & 78.3 \\
PVT-Large\cite{Wang_2021_ICCV}& Transformers  & $\greencheckmark$ & 61.4 & \textbf{9.8} & 81.7\\
ViT-Base16\cite{dosovitskiy2020image} & Transformers  & \redxmark & 86.6& 16.0 & 81.8\\
T2T-ViT-24\cite{yuan2021tokenstotoken}& Transformers & $\greencheckmark$ & 63.9 & 13.2 & 82.2 \\
XCiT-M24\cite{ali2021xcit} & Linear Transformers  & \redxmark & 84.0 & 16.2 & 82.7\\ 
ResNeXt101-64x4d\cite{xie2017aggregated}& ConvNets  & $\greencheckmark$ & 84.0 & 16.0 & 82.9 \\
TNT-B\cite{han2021transformer}& Transformers & \redxmark & 65.6 & 14.1 & 82.9 \\
RegNetY-16G\cite{xu2021regnet}& ConvNets  & $\greencheckmark$ & 84.0 & 16.0 & 82.9 \\
SOFT\cite{lu2021soft}& Linear Transformers  & $\greencheckmark$ & 64.0 & 11.0 & 83.1 \\ 
Swin-B\cite{liu2021swin}& Transformers  & $\greencheckmark$ & 88.0 & 15.4 & 83.3\\
Twins-SVT-L\cite{chu2021twins}& Hybrid  & $\greencheckmark$ & 99.2 & 14.8 & 83.7 \\
PVTv2-B5\cite{wang2022pvtv2}& Transformers  & $\greencheckmark$ & 82.0 & 11.8 & 83.8 \\
Focal-Base\cite{yang2021focal} & Transformers  & $\greencheckmark$ & 89.8 & 16.0 & 83.8 \\
DAT-B\cite{xia2022vision} & Transformer & \greencheckmark & 88.0 & 15.8 & 84.0 \\
Quadtree-B-b4 \cite{tang2022quadtree}& Linear Transformers   & $\greencheckmark$ & 64.2 & 11.5 & 84.0 \\
VVT-L(ours)& Linear Transformers  & $\greencheckmark$ & 61.8 & 10.8 &  \textbf{84.1} \\
\hline
\end{tabular}
\label{table:imagenet1k}
\end{table*}

\begin{table}[h]
\tabcolsep=0.25cm
\centering
\caption{\textbf{CIFAR-10~\cite{krizhevsky2009learning} classification results.}}
\begin{tabular}{r|c|c|c}
Method & Style & MS out & Top-1 Acc(\%)  \\
\hline
ViT-Small~\cite{dosovitskiy2020image} & Transformers & \redxmark & 81.9 \\
TNT-S\cite{han2021transformer}& Transformers & \redxmark & 85.8 \\
Swin-T\cite{liu2021swin}& Transformers  & $\greencheckmark$ & 89.8\\
PVT-Small\cite{Wang_2021_ICCV}& Transformers  & $\greencheckmark$ & 91.1 \\
ResNet50\cite{he2016deep}& ConvNets  & $\greencheckmark$  & 92.5 \\
Twins-SVT-S\cite{chu2021twins}& Hybrid  & $\greencheckmark$&  93.8  \\
PVTv2-B2\cite{wang2022pvtv2}& Transformers  & $\greencheckmark$ & 95.7 \\
VVT-S(ours)& Linear Transformers  & $\greencheckmark$ & \textbf{96.1} \\
\hline
\end{tabular}
\label{table:Cifar10}
\end{table}

\begin{table}[h]
\tabcolsep=0.25cm
\centering
\caption{\textbf{CIFAR-100~\cite{krizhevsky2009learning} classification results.}}
\begin{tabular}{r|c|c|c}
Method & Style & MS out & Top-1 Acc(\%)  \\
\hline
ViT-Small~\cite{dosovitskiy2020image} & Transformers & \redxmark & 47.4 \\
TNT-S\cite{han2021transformer}& Transformers & \redxmark & 71.5 \\
Swin-T\cite{liu2021swin}& Transformers  & $\greencheckmark$ & 74.8\\
PVT-Small\cite{Wang_2021_ICCV}& Transformers  & $\greencheckmark$ & 76.3 \\
ResNet50\cite{he2016deep}& ConvNets  & $\greencheckmark$  & 79.2 \\
Twins-SVT-S\cite{chu2021twins}& Hybrid  & $\greencheckmark$&  81.6 \\
PVTv2-B2\cite{wang2022pvtv2}& Transformers  & $\greencheckmark$ & 83.9 \\
VVT-S(ours)& Linear Transformers  & $\greencheckmark$ & \textbf{84.6} \\
\hline
\end{tabular}
\label{table:Cifar100}
\end{table}

\begin{table*}[t]
%\scriptsize
% \tabcolsep=0.16cm
\setlength{\tabcolsep}{0.6cm}
\centering 
\caption{\textbf{Semantic segmentation results on ADE20K\cite{zhou2017scene} validation set. }
% We follow the 
Competing results come from PVTv2~\cite{wang2022pvtv2} and Twins~\cite{chu2021twins}. All backbone networks are pre-trained on ImageNet-1k\cite{deng2009imagenet}.}
\begin{tabular}{r|ccc | ccc}
\multirow{2}{*}{Method} & \multicolumn{3}{c}{Semantic FPN (PVT~\cite{Wang_2021_ICCV} setting)} & \multicolumn{3}{|c}{UPerNet (Swin~\cite{liu2021swin} setting)} \\ 
\cline{2-7} 
                          & Params    & GFLOPs   & mIoU(\%) & Params    & GFLOPs   & mIoU(\%)  \\ \hline
ResNet50    \cite{he2016deep}              &    28.5       & 45.6         &      36.7 &-&-&- \\
PVT-Small\cite{Wang_2021_ICCV}                  & \textbf{28.2}           &    44.5      &     39.8 &-&-&-   \\
Swin-T\cite{liu2021swin} & 31.9 & 46.0 & 41.5  & 59.9 & 237 & 44.5\\ 
Twins-SVT-S\cite{chu2021twins} & 28.3 & \textbf{37.0} & 43.2 & \textbf{54.4} & \textbf{228} & 46.2 \\
PVTv2-b2\cite{wang2022pvtv2} & 29.1 &45.8 & 45.2 &-&-&- \\
VVT-S(ours)                      &    29.2       & 50.9         &  \textbf{45.6} &
55.5&240 & \textbf{46.8}          \\ \hline
ResNet101\cite{he2016deep}                  &    47.5       &    65.1      &     38.8  &-&-&- \\
ResNeXt101-32x4d\cite{xie2017aggregated}           &        \textbf{47.1}   &    64.7      &     39.7 &-&-&-   \\
PVT-Medium\cite{Wang_2021_ICCV}                 &     51.6      &    \textbf{61.0}      &      41.6&-&-&-  \\
SWIN-S\cite{liu2021swin} & 53.2 & 70.0 & 45.2 & 81.3 & 261 & 47.6 \\
Twins-SVT-B\cite{chu2021twins} & 60.4 & 67.0 & 45.3 & 88.5 & 261 & 47.7 \\
PVTv2-b3\cite{wang2022pvtv2} & 49.0& 62.4& 47.3 &-&-&- \\
VVT-M(ours)                  &   51.7        &   70.8       & \textbf{47.4}  & \textbf{77.9} & \textbf{260} &  \textbf{48.1}        \\ \hline
ResNeXt101-64x4d \cite{xie2017aggregated}           &     86.4      &     103.9     &  40.2&-&-&-      \\
PVT-Large\cite{Wang_2021_ICCV}                  &    \textbf{65.1}       &    79.6      & 42.1&-&-&-       \\
SWIN-B\cite{liu2021swin} & 91.2 & 107.0 & 46.0 &121&299&48.1   \\
Twins-SVT-L\cite{chu2021twins} & 103.7 & 102.0 & 46.7 &133 & 297 & 48.8 \\
PVTv2-b4\cite{wang2022pvtv2} & 66.3& 81.3 & 47.9 &-&-&-  \\
VVT-L(ours)                   &   65.5        &   \textbf{78.1}       &    \textbf{47.9} & \textbf{91.9} & \textbf{267} &  \textbf{48.8}      \\ \hline
\end{tabular}
\label{table: segmentation}
\end{table*}

\section{Experiments}
To verify the effectiveness of our method, we conduct extensive experiments on the CIFAR-100~\cite{krizhevsky2009learning} and ImageNet-1k~\cite{deng2009imagenet} datasets for image classification, and on the ADE20K~\cite{zhou2017scene} dataset for semantic segmentation.
Specifically, we first make a comparison with existing state-of-the-art methods, and then give an ablation study over the design of VVT.
\subsection{Image Classification}

\noindent\textbf{ImageNet-1k}
The image classification experiments are conducted on the ImageNet-1k~\cite{deng2009imagenet} dataset, which contains 1.28 million training images and 50 thousand validation images from $1,000$ categories.
We follow the training hyper-parameters of \cite{Wang_2021_ICCV}.
In detail, the model is trained using an AdamW~\cite{kingma2015adam} optimizer with a weight decay of $0.05$ and a momentum of $0.9$. 
We use an initial learning rate of $5\times 10^{-4}$ and decrease it by a cosine schedule~\cite{loshchilov2017sgdr}, with $5$ epochs for warming up.
We also adopt the same data augmentation strategy as in~\cite{Wang_2021_ICCV}, including random cropping, random horizontal flipping, \etc~
All the models are trained on the training set for $300$ epochs with a crop size of $224\times224$.
We use the top-1 accuracy as the evaluation metric.

The quantitative results are shown in Table \ref{table:imagenet1k}, which cover the popular transformer-based and CNN-based classification networks. 
For fair comparison, we compare the VVT variants respectively with the networks using similar parameters, split by solid lines.
For example, VVT-L performs better than PVTv2-B5 \cite{Wang_2021_ICCV} but only uses around 70\% parameters of the latter.
In particular, our medium-size variant VVT-M surpasses  the large models Twins-SVT-L~\cite{chu2021twins} and Swin-B~\cite{liu2021swin} with substantially fewer parameters (51.7\% and 45.5\% respectively).
Moreover, compared to the state-of-the-art CNN-based networks such as RegNet \cite{radosavovic2020designing} and ResNeXt \cite{xie2017aggregated}, VVT shows stronger performance while using a similar number of parameters.

It is worth noting that VVT directly computes self-attention on the high-resolution feature maps, while the competitors such as PVT~\cite{Wang_2021_ICCV}, Swin Transformer~\cite{liu2021swin}, Twins~\cite{chu2021twins} and SOFT~\cite{lu2021soft} rely on  sub-sampling or window self-attention to reduce computational cost.
The GFLOPs of VVT could be further reduced if similar subsampling or windowing was used, we validate this in Fig.~\ref{fig:memory}. \\

\noindent\textbf{CIFAR-10 and CIFAR-100.}
It is known that vision transformers may suffer critical performance drops on small-scale datasets.
To validate the performance of our method on small-scale datasets, we use well-known CIFAR-10 and CIFAR-100~\cite{krizhevsky2009learning} as target datasets.
Tables~\ref{table:Cifar10} and~\ref{table:Cifar100} show our VVT-S and competitors' results. 
All models are trained from scratch using an AdamW optimizer with a weight decay of 0.05 and a momentum of 0.9. We use an initial learning rate of $5\times 10^{-4}$ and train for 300 epochs. All models are trained and tested on $224 \times 224$
resolution for fair comparison. 
We see that our VVT outperforms all competing methods on both CIFAR-10 and CIFAR-100.
We hypothesize that for small datasets, stronger inductive bias may make the model more data efficient.
We can see ViT does not have inductive bias and performs poorly.
Other models that have different types of locality bias achieve better results than ViT. 
Our VVT introduces self-attention with 2D locality and achieves the best results.

\begin{figure*}[h]
% \centering
   \begin{center}
   {\includegraphics[width=1    \linewidth]{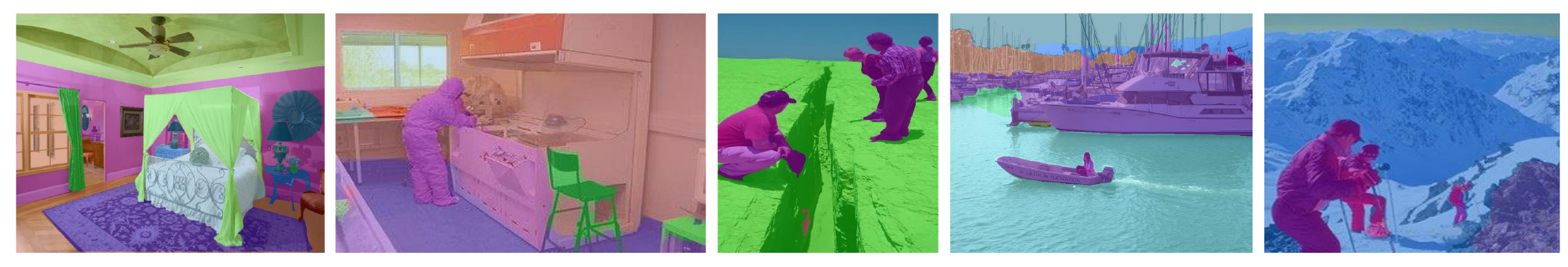}} 
   \end{center}
   \vspace{-5mm}
\caption{Qualitative results of semantic segmentation on ADE20K\cite{zhou2017scene}. The results are generated by VVT-L based Semantic FPN\cite{kirillov2019panoptic}.}
\label{fig:results}
\end{figure*}

\subsection{Semantic Segmentation}
We utilize the challenging ADE20K~\cite{zhou2017scene} dataset to evaluate our idea in semantic segmentation.
It has $150$ semantic classes, with $20{,}210$, $2{,}000$ and $3{,}352$ images for training, validation and testing. 
With the VVT models (pre-trained on ImageNet-1k) as the backbone, we provide the results using two segmentation methods in Table~\ref{table: segmentation} and show qualitative samples in Fig.~\ref{fig:results}. 
Specifically, we pick Semantic FPN~\cite{kirillov2019panoptic} and UPerNet~\cite{xiao2018unified} as the semantic segmentation architecture, respectively following the settings of PVT~\cite{Wang_2021_ICCV} and Swin Transformer~\cite{liu2021swin}.
The quantitative results show that our method is consistently superior to the competing CNN-based methods.
For example, using Semantic FPN, VVT-S outperforms ResNet50~\cite{he2016deep} by 8.9\%, and VVT-L is 7.7\% better than ResNeXt101-32x4d~\cite{xie2017aggregated}, using mIoU as the metric.
VVT also shows a more favourable segmentation result than the transformer-based competitors such as the Swin Transformer. 
A similar phenomenon is observed if taking UPerNet~\cite{xiao2018unified} as the segmentation method.
These results validate that the extracted features of VVT are also valuable for semantic segmentation, which benefits from the locality mechanism together with the global attention.

\subsection{Ablation Study}
\begin{figure}[t!]
   \begin{center}
   {\includegraphics[width=1\linewidth]{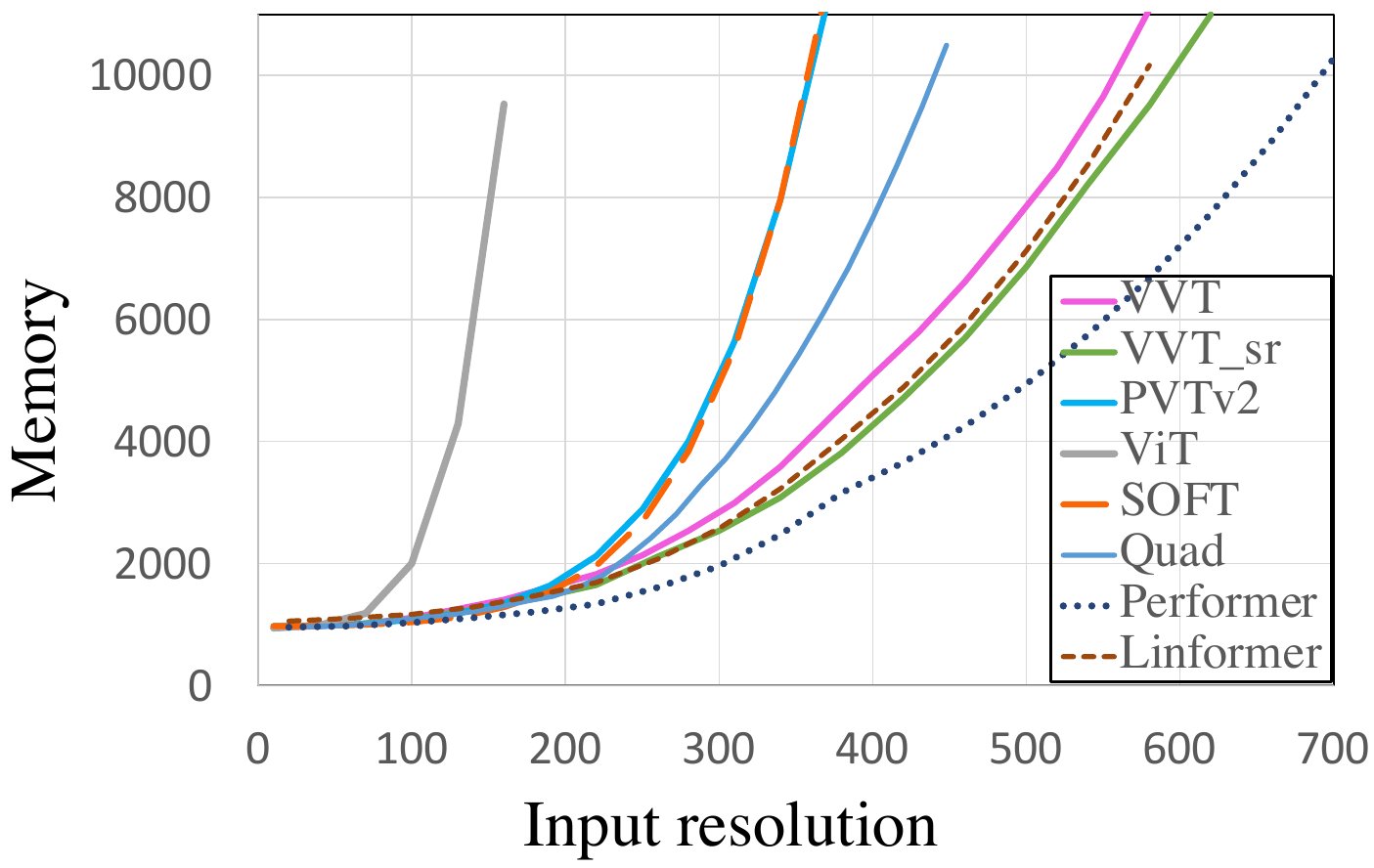}}
   \end{center}
\caption{Comparison of attention memory footprints growth rates between VVT and recent efficient transformer methods: PVTv2\cite{wang2022pvtv2}, SOFT\cite{lu2021soft}, Quadtree\cite{tang2022quadtree}, Performer\cite{choromanski2020rethinking} and Linformer\cite{wang2020linformer}. 
We set the feature dimension to 128 and the number of heads to 1 for all methods.         
Following their official implementation, the spatial reduction ratio is set as 8 on PVTv2 and SOFT, and 4 on Quadtree.
For VVT\_sr, we integrate a spatial reduction ratio of 4 into our VVT, which further improves efficiency.
}
\label{fig:memory}
\end{figure}

{\textbf{Computational Overhead.}}
We plot the GFLOPs growth rates with respect to the input image size on the right of Fig.~\ref{fig:intro}.
The growth rate of VVT (purple) is substantially lower than other transformer-based methods, and even slightly lower than ResNet~\cite{he2016deep}.
However, theoretical GFLOPs may arguably not reflect real computational overhead. 
To further demonstrate the computational efficiency of Vicinity Attention, we compare GPU memory footprints of Vicinity Attention against recent efficient vision transformer methods\cite{wang2022pvtv2, lu2021soft, tang2022quadtree,choromanski2020rethinking,wang2020linformer} in Fig.~\ref{fig:memory}.
All results are obtained under the same settings including feature dimension, number of heads \etc\
Note that although VVT, SOFT~\cite{lu2021soft}, Performer~\cite{choromanski2020rethinking}, Quad~\cite{tang2022quadtree} and Linformer~\cite{wang2020linformer} all have linear complexity, their actual memory consumption may vary according to their specific algorithms.

From Fig.~\ref{fig:memory}, we can make the following observations: 
(i) Our method consumes relatively more memory than Performer\cite{choromanski2020rethinking} and Linformer\cite{wang2020linformer}, as both methods are designed for 1D sequences and show inferior performance on vision tasks.
(ii) Compared to efficient vision transformers, \ie, PVTv2\cite{wang2022pvtv2}, SOFT\cite{lu2021soft} and Quadtree\cite{tang2022quadtree},
the memory consumption of Vicinity Attention grows notably slower, allowing inputs with much higher resolutions.
(iii) The spatial reduction (sr) is a computation reduction technique proposed by PVT~\cite{wang_PVT_2021}. 
Specifically, it reduces the spatial dimensions of the key and value vectors in the self-attention module with convolution or pooling operations, thus the complexity of the self-attention computation is reduced.
PVTv2\cite{wang2022pvtv2}, SOFT\cite{lu2021soft}, and Quadtree\cite{tang2022quadtree} all adopt the sr strategy to reduce computation, while the standard VVT directly calculates attention on the original sequence without sr. 
If we also integrate sr, the
VVT\_sr in Fig.~\ref{fig:memory} validates that the efficiency can be further improved with a spatial reduction ratio of 4.
Finally, in Table.~\ref{Table: memory number},
we show memory footprints of the models with various attention modules that correspond to Table~\ref{ablation table:attention}.
In summary, this ablation study validates that VVT successfully mitigates the computational issue in the vision transformers.\\

\begin{table}[t]
\setlength{\tabcolsep}{0.5cm}
\centering 
\caption{\textbf{Analysis of locality constraint on CIFAR-100\cite{krizhevsky2009learning} and ImageNet-1k\cite{deng2009imagenet}.} 
}
\begin{tabular}{llc}
\hline
Datasets &Method & Top-1 Acc(\%) \\
\midrule
\multirow{3}{*}{{CIFAR-100}}
&VVT w/o locality &  82.13 \\
&VVT 1D locality & 76.78 \\
&VVT 2D locality  & 82.92  \\
\midrule
\multirow{3}{*}{{ImageNet-1k}}
&VVT w/o locality  &78.0  \\
&VVT 1D locality  & \xmark \\
&VVT 2D locality  &79.4  \\
\hline
\end{tabular}
\label{table: ablate locality}
\end{table}

\begin{figure*}[t!]
   \begin{center}
   {\includegraphics[width=0.75\linewidth, height=5cm]{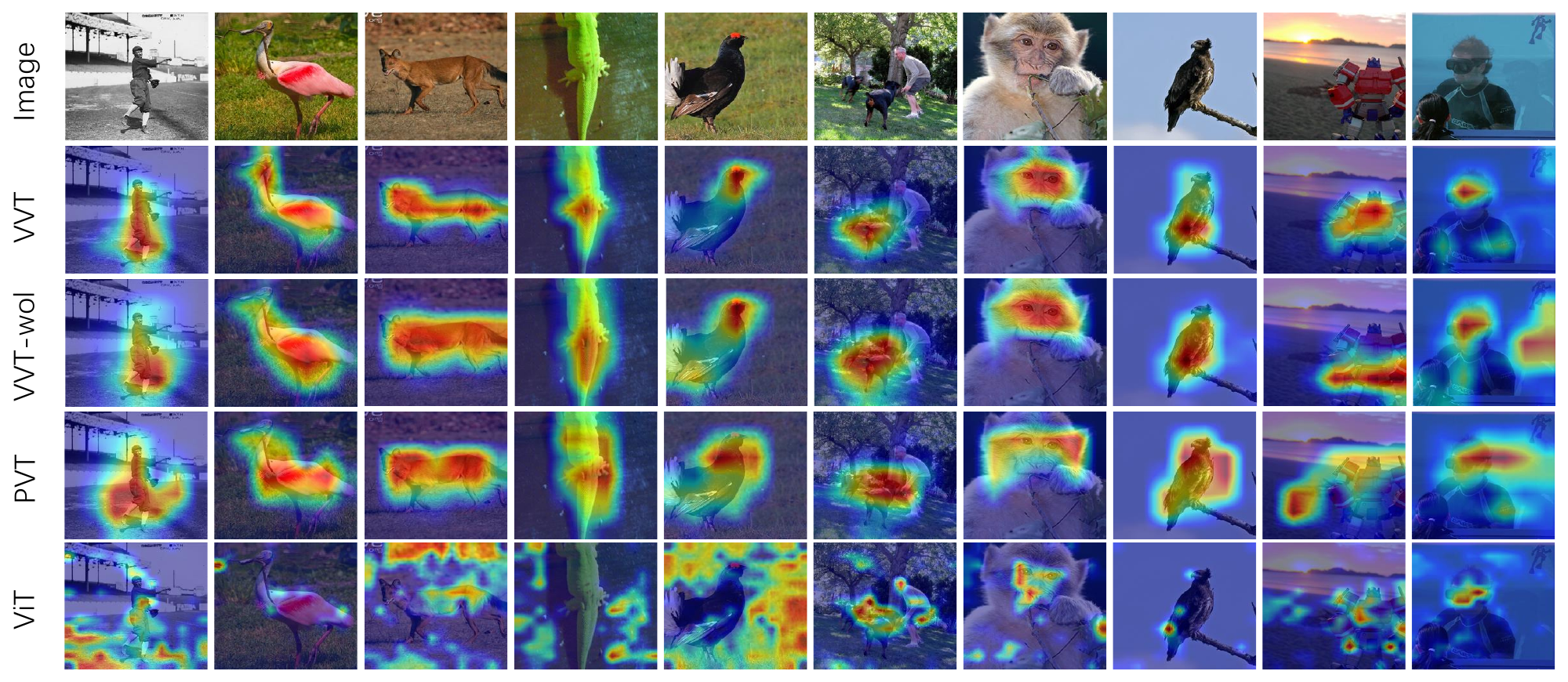}}
   \end{center}
\caption{We compare the class-wise attention maps of VVT against VVT w/o locality (VVT-wol in the figure),  PVT\cite{Wang_2021_ICCV} and  ViT\cite{dosovitskiy2020image}. The attention maps are generated using Grad\_CAM\cite{selvaraju2017grad}. We observe that the activation of PVT and VVT w/o locality is more concentrated than ViT due to the fact that overlapping patch embedding or pyramid structure may also introduce locality.
In VVT, we enforce 2D locality bias in terms of the attention module, so that the activation areas are more concentrated and appropriately located on the object.}
\label{fig:cam2}
\end{figure*}

\noindent{\textbf{Effect of Locality Constraint.}}
The \textit{Vicinity} locality mechanism is introduced in the self-attention module of VVT.
We show its effectiveness on the CIFAR-100~\cite{krizhevsky2009learning} and the ImageNet-1k~\cite{deng2009imagenet} datasets in Table~\ref{table: ablate locality}.
Experiments are performed on the VVT-Tiny structure.
We compare VVT with a variation without locality (denoted as VVT w/o locality) and the one with 1D locality based on cosFormer~\cite{zhen2022cosformer} (denoted as VVT 1D locality). 
As shown in the table, our VVT achieves the best performance among all the others on the CIFAR-100 dataset.
By comparing the VVT w/o locality and VVT 1D locality, we find that if we wrongly enforce the locality bias, the performance will drop critically. 
To further demonstrate its effectiveness on large-scale data, we also conduct the same ablation on ImageNet-1k.
We observe that the model fails to converge with 1D locality which indicates that the 1D cosFormer cannot be trivially transplanted to 2D vision data. 
Finally, our VVT outperforms the VVT w/o locality by $1.2\%$, showing the efficiency of our linear locality attention.

Furthermore, we show the qualitative examples of Grad-CAM~\cite{selvaraju2017grad} from VVT and the competitors in Fig.~\ref{fig:cam2}. 
As shown, the locality mechanism encourages the tokens to assign higher attention to 2D neighbours, and hence the class-wise activations are more concentrated and accurately located on the target object regions of Grad-CAM.
These results validate that the introduction of the 2D locality mechanism  helps the VVT models generate more reliable object features, which tend to be beneficial in image classification and downstream tasks.\\

\begin{table}[t!]
\setlength{\tabcolsep}{0.15cm}
        \centering
        \caption{\textbf{Ablation of different attention choices.} We fix the architecture to VVT-T and test different attention modules on CIFAR-100 and ImageNet-1k. cosFormer fails to converge under the same setting on ImageNet1K.}
        \begin{tabular}{cllccc}
\hline
Dataset & Architecture & Attention  & Memory & Top-1 Acc($\%$)  \\
\hline
\multirow{7}{*}{{CIFAR-100}}
&VVT-T & GCNet\cite{Cao_2019_ICCV}  &  & 66.34 \\
&VVT-T & cosFormer\cite{zhen2022cosformer}  &  & 76.78 \\
&VVT-T & Performer\cite{choromanski2020rethinking}   & & 79.86 \\
&VVT-T & Linformer\cite{wang2020linformer}  & & 80.95 \\
&VVT-T & PVTv2\cite{wang2022pvtv2}   & & 82.16 \\
&VVT-T & Vanilla\cite{vaswani2017attention}   & & 82.68 \\
&VVT-T & Ours   & & 82.92 \\
\hline
\multirow{7}{*}{{ImageNet-1k}}
&VVT-T & GCNet\cite{Cao_2019_ICCV}  & & 76.6 \\
&VVT-T & cosFormer\cite{zhen2022cosformer}  & & \xmark \\
&VVT-T & Performer\cite{choromanski2020rethinking}  & &77.2 \\
&VVT-T & Linformer\cite{wang2020linformer}  & &79.1 \\
&VVT-T & PVTv2\cite{wang2022pvtv2}   & &79.3 \\
&VVT-T & Vanilla\cite{vaswani2017attention}   & & 80.0 \\
&VVT-T & Ours  &  & 79.4 \\
\hline
\end{tabular}
\label{ablation table:attention}
\end{table}

\begin{table}[]
\setlength{\tabcolsep}{0.1cm}
\caption{Quantitative memory footprints of networks in table~\ref{ablation table:attention}. All models have the same network structure except the attention module. All results are obtained with a batch size of 16 on a GPU with 32GB VRAM.}
\centering\scalebox{0.92}{
\begin{tabular}{c|cccccc|c}
\hline
Resolutions & Vanilla & GCNet & cosFormer & Performer & Linformer & PVTv2  & VVT  \\ \hline
$224^2$     & 4.8     & 1.8   & 3.1       & 2.5       & 3.0       & 2.4  & 3.2  \\
$384^2$     & 29.7    & 5.2   & 8.8       & 7.4       & 8.7      & 9.1  & 9.2  \\
$512^2$     & OOM     & 9.1   & 15.7      & 13.0      & 15.4      & 20.8 & 16.1 \\ \hline
\end{tabular}}
\label{Table: memory number}
\end{table}

\noindent
\noindent{\textbf{Comparison with Existing Attention Modules}}
We compare with several existing self-attention methods: cosFormer\cite{zhen2022cosformer}, Performer\cite{choromanski2020rethinking}, Linformer\cite{wang2020linformer} PVTv2\cite{wang2022pvtv2} and vanilla self-attention~\cite{vaswani2017attention} .
For all methods, we adopt the VVT-Tiny network structure and only replace the attention
block with competing attention modules for a fair comparison. 
All experiments are conducted using the same network configuration and training setting. 
We report results on the CIFAR-100 and the ImageNet1K in Table \ref{ablation table:attention}.
As shown, on CIFAR-100 our Vicinity attention outperforms both the vanilla attention and all the alternative efficient methods.
On ImageNet1K, we observe that cosFormer\cite{zhen2022cosformer} fails to converge due to the false 1D locality. 
Compared to other efficient methods including GCNet~\cite{Cao_2019_ICCV}, Performer~\cite{choromanski2020rethinking}, Linformer~\cite{wang2020linformer} and PVTv2~\cite{wang2022pvtv2}, VVT still achieves better classification result.
Further, vanilla softmax attention~\cite{dosovitskiy2020image} is $0.6\%$ higher than our VVT but at a cost of substantially higher computational complexity (see Table.~\ref{Table: memory number}), which makes it implausible to extend to any larger models with pyramid structures.
In Table.~\ref{Table: memory number}, we display the quantitative memory footprints of the aforementioned models in Table~\ref{ablation table:attention}. 
All models share the same network structure except the attention module with a batch size of 16, 
allowing for a fair comparison of different attention mechanisms. 
Our VVT has linear complexity and similar memory footprints to existing linear attention methods, \ie, cosFormer~\cite{zhen2022cosformer}, Linformer~\cite{wang2020linformer} and Performer~\cite{choromanski2020rethinking}, whereas our performances on both datasets are superior.
Finally, VVT has better memory efficiency than PVTv2~\cite{wang2022pvtv2} and vanilla softmax attention~\cite{dosovitskiy2020image}. \\

\begin{figure}[t!]
   \begin{center}
   {\includegraphics[width=0.9\linewidth]{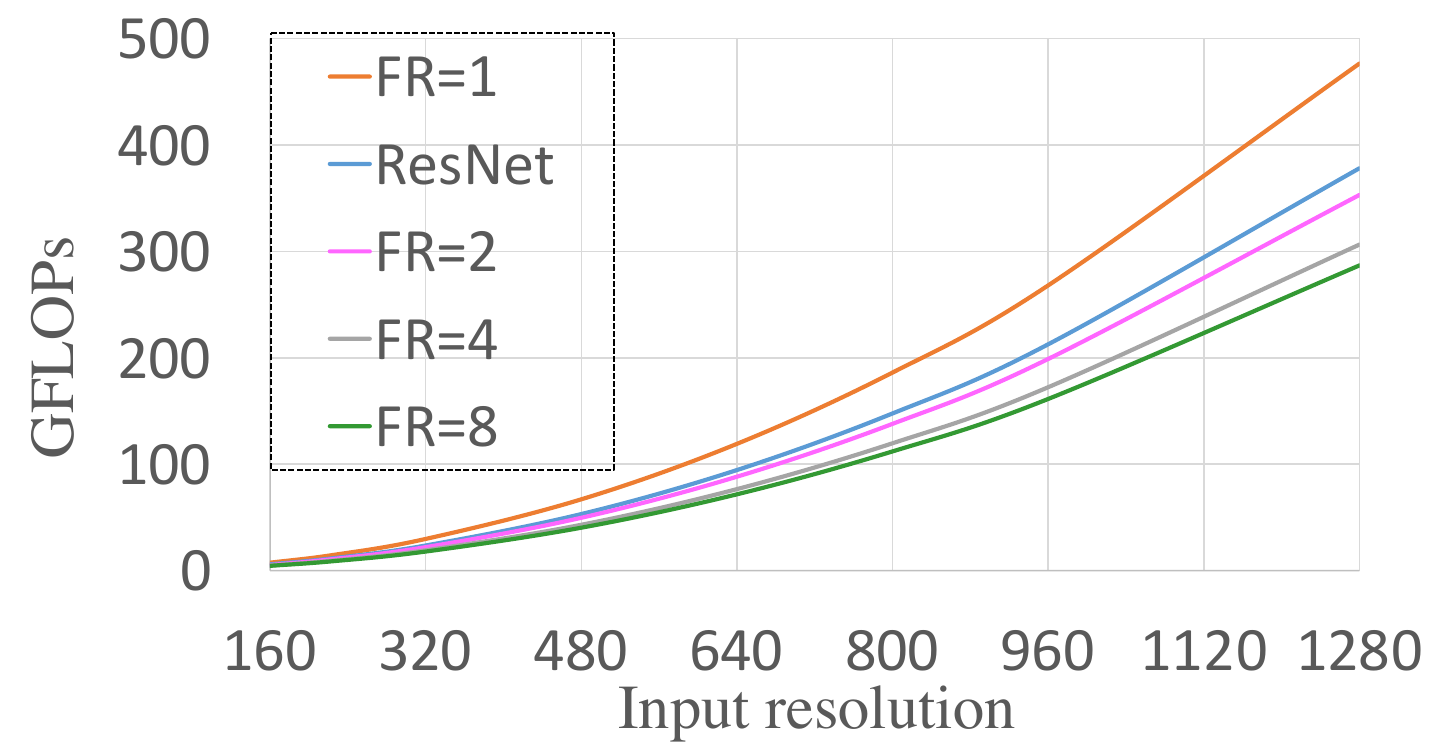}}
   \end{center}
\caption{GFLOPs of VVT using different feature reduction rates. When FR ratio is increased larger than 4, the growth ratio tends to be saturated.}
\label{fig: fr GFLOPs}
\end{figure}

\begin{table}[t]
  \centering
  \setlength{\tabcolsep}{0.5cm}
  \caption{\textbf{Results of VVT on CIFAR100 using different feature reduction ratio (FR).} We empirically find that the $\text{FR}=2$ leads to the best performance.}
  \begin{tabular}{ c|c|c } 
  \hline
  Method& Feature Reduction ratio  &  Top-1 Acc(\%)  \\
  \hline
PVT\cite{Wang_2021_ICCV} & -- & 81.67\\
VVT &FR=1 &  82.89\\
VVT &FR=2 & 82.92\\
VVT &FR=4 & 82.00\\
VVT &FR=8 & 80.60\\
  \hline
  \end{tabular}
  \label{table: ablation fr}
\end{table}

\noindent{\textbf{Analysis of Vicinity Attention Block.}}
In this section, we investigate the proposed Vicinity Attention Block from three perspectives:
(1) we show that feature reduction attention (FRA) can significantly reduce computation without harming performance; (2) We validate that feature preserving connection (FPC) is effective to strengthen feature extraction ability; 
and (3) we report the effectiveness of FRA and FPC on an existing linear attention method. 
In summary, we demonstrate that FRA effectively reduces computational complexity and FPC preserves representation ability without causing degeneration of performance.

By FRA, we reduce the feature dimension to fulfill the assumption of $N\gg d$, which forces the computational complexity to approach linear.
In Fig.~\ref{fig: fr GFLOPs}, we ablate the computational cost under different feature reduction (FR) ratios.
We can observe an obvious GFLOPs decrease when increasing the FR ratio from 1 to 2 (reducing the feature dimension by half).
However, when the FR ratio further increases, the reduction tends to saturate. 
This is because the computational cost of the self-attention module is already significantly reduced and the remaining model parts dominate the computational overhead.
Additionally, we ablate the performance under different FR ratios on the CIFAR-100 dataset. 
As shown in Table~\ref{table: ablation fr}. 
When the FR ratio is small, the model retains a similar performance (\ie, 82.89\% vs.~82.92\%), indicating that the result is not affected.
When the FR ratio increases to a number like 8, a clear performance drop is observed.
As a trade-off, we set the FR ratio as 2 for our formal model settings.

\begin{table}[htb]
  \centering
\setlength{\tabcolsep}{0.9cm}
\caption{\textbf{Results of VVT-tiny with and without feature preserving connection(FPC) on ImageNet-1k.}}
  \begin{tabular}{ c|c|c } 
  \hline
  Method& FPC  & Top-1 Acc(\%)  \\
  \hline
VVT & \redxmark       & 77.9\\
VVT & \greencheckmark & 79.4\\
  \hline
  \end{tabular}
  \label{table: ablation fp}
\end{table}

\begin{table}[htb]
  \centering
\setlength{\tabcolsep}{0.67cm}
\caption{\textbf{Results of integrating feature reduction attention (FRA) and feature preserving connection (FPC) with Performer\cite{choromanski2020rethinking} on ImageNet-1k.}}
  \begin{tabular}{ c|c|c } 
  \hline
  Method & FRA\&FPC  & Top-1 Acc(\%)  \\
  \hline
Performer\cite{choromanski2020rethinking} & \redxmark       & 77.2\\
Performer\cite{choromanski2020rethinking} & \greencheckmark & 77.3\\
  \hline
  \end{tabular}
  \label{table: performer vicinity attention block}
\end{table}

\begin{table}[t]
\setlength{\tabcolsep}{0.25cm}
\centering 
\caption{\textbf{Classification results on ImageNet-1k under the input scale 384 $\times$ 384.}
}
\begin{tabular}{ccccc}
\hline
Method & Resolution & Param & GFLOPs &Top-1 Acc(\%) \\
\midrule
VVT-T  & $384^2$ & 12.9  & 8.8 & 80.3\\
\midrule
Swin-T\cite{liu2021swin} & $384^2$ & 29.0 &  14.0    &   82.2 \\
CVT-21\cite{wu2021cvt} &$384^2$& 32.0 & 24.9 & 83.3 \\
VVT-S & $384^2$ & 25.5 & 16.5 & 83.4 \\
\midrule
Swin-S\cite{liu2021swin} & $384^2$& 50.0 & 27.0 & 83.9 \\
VVT-M  & $384^2$& 47.9    & 27.7 & 84.5 \\ 
\midrule
ViT-Base16\cite{dosovitskiy2020image} &$384^2$& 86.6 & 55.5    & 77.9 \\ 
Swin-B\cite{liu2021swin}& $384^2$ & 88.0 & 47.0 & 84.5 \\
VVT-L & $384^2$&  61.8 & 31.8 &  84.7\\
\hline
\end{tabular}
\label{table: ablate resolution}
\end{table}

In the Vicinity Attention Block, FPC is added to compensate for reduced feature dimensions. 
Ablation of FPC is reported in Table \ref{table: ablation fp}. 
The results indicate that FPC effectively retains the original feature distribution, and 
improves the feature extraction ability of the Vicinity attention module even when the feature dimension is reduced.

To further demonstrate the effectiveness of Vicinity Attention Block, we integrate FRA and FPC with an existing linear attention method, \ie, Performer\cite{choromanski2020rethinking}. 
As shown in Table \ref{table: performer vicinity attention block}, we attach the Performer attention block on VVT-T as a baseline.
After integrating FRA and FPC, the computation is reduced while we can still observe an accuracy improvement (\ie, 77.2\% vs.~77.3\%). 
It validates that our FRA and FPC yield performance gains in the existing linear attention approach.

\noindent{\textbf{Image Classification with Different Input Size}}
Table \ref{table: ablate resolution} lists the  performance of VVT with higher input resolution i.e., 384. 
Obviously, larger input resolution leads to better top-1 accuracy but requires larger computation. Compared to competing methods such as ViT\cite{dosovitskiy2020image}, Swin\cite{liu2021swin} and CVT\cite{wu2021cvt}, VVT achieves better results with fewer parameters and 
a comparable computational overhead. \\

\begin{figure}[htb]
   \begin{center}
   {\includegraphics[width=0.9\linewidth]{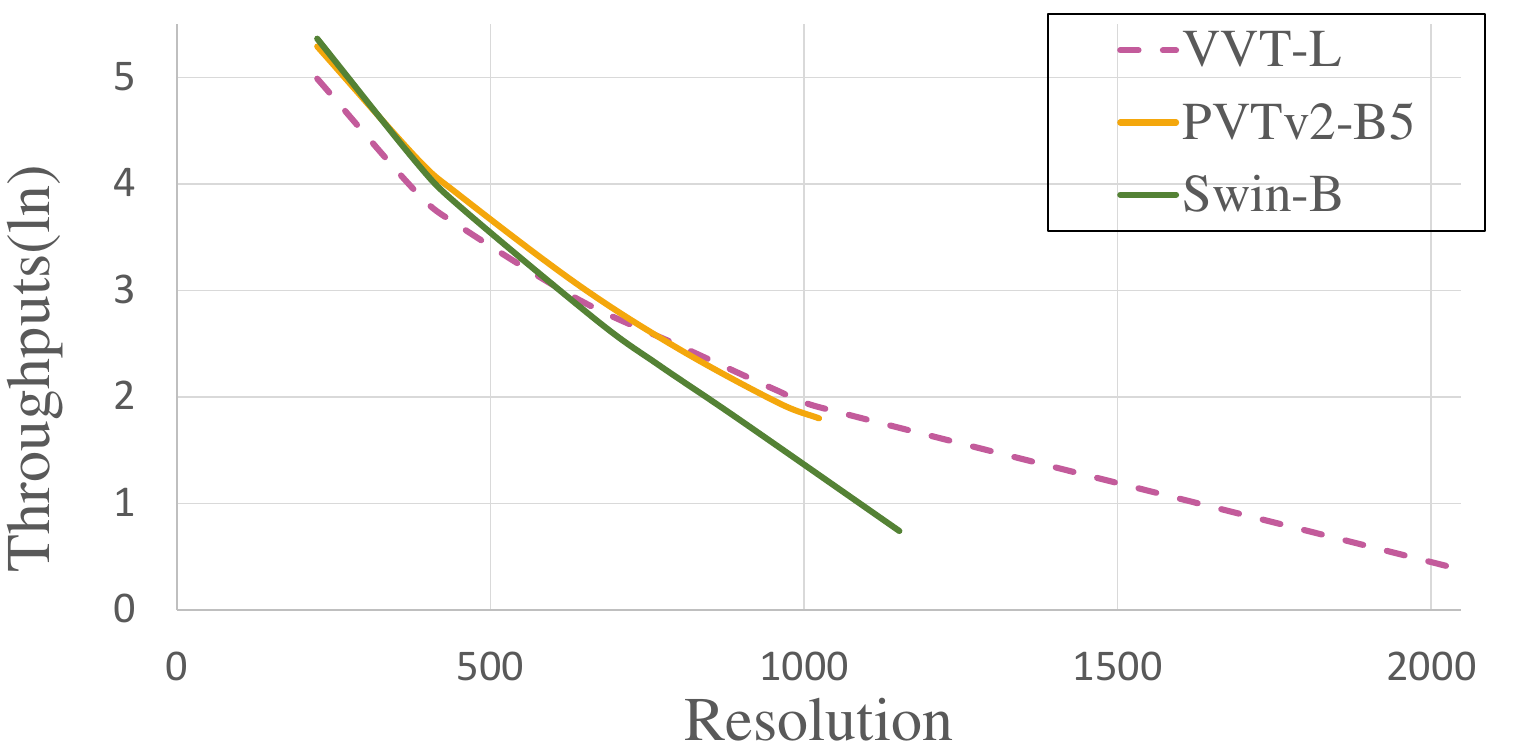}}
   \end{center}
\caption{Throughput-resolution curve. We take $log$ of throughput for better readability since throughput differences saturate in the high-resolution regime.}
\label{fig: through}
\end{figure}

\noindent\textbf{Analysis of Overlapping Patch Embedding and Convolutional Feed-Forward.}
In standard VVT, we adopt the
overlapping patch embedding module and Convolutional Feed-Forward proposed by Wang \etal~\cite{wang2022pvtv2}.
Both modules may also contribute to locality bias and improve results~\cite{wu2021cvt}. 
Table~\ref{table: overlapping patch embed} ablates our Vicinity Attention without these two modules, compared to existing methods under the same setting. 
As shown, we observe that without the contribution of the overlapping patch embedding and convolutional feed-forward, VVT also outperforms competing methods in both CIFAR-10 and ImageNet-1k which shows the efficacy of our attention method.

\begin{table}[htb]
  \centering
\setlength{\tabcolsep}{0.15cm}
\caption{\textbf{Results without Overlapping Patch embedding and Convolutional Feed-Forward. VVT-Tiny* indicates that we adopt the VVT-Tiny but use the vanilla patch embedding and Feed-Forward.}}
  \begin{tabular}{ c|c|c } 
  \hline
  Method& CIFAR-10 Acc(\%)  & ImagNet Acc(\%)   \\
  \hline
Deit-Tiny~\cite{touvron2021training} & 81.9 &  72.2 \\
PVT-Tiny ~\cite{wang_PVT_2021} & 88.9  &    75.1  \\
VVT-Tiny* & 89.5 &    75.3  \\
  \hline
  \end{tabular}
  \label{table: overlapping patch embed}
\end{table}

\noindent{\textbf{Throughput}}
Fig.~\ref{fig: through} illustrates actual image throughputs of the proposed VVT compared to PVTv2\cite{wang2022pvtv2} and Swin transformer\cite{liu2021swin}. We run the test on one local 2080 Ti GPU with 10 Gb memory.
Since the theoretical computational complexity of Vicinity attention is $O(N(2d)^2 + 2d^2)$,
in low resolution regime where sequence length $N$ does not dominate the complexity,
VVT shows a relatively slow speed compared to PVTv2\cite{wang2022pvtv2} and Swin\cite{liu2021swin}.
When the input resolution is gradually increased, the VVT starts demonstrating better inference speed, while competing methods exhaust GPU memory in the high-resolution regime. 
In summary, due to the better overall efficiency curve we consider slightly increased time consumption in the low-resolution regime is a price worth paying for our better accuracy.

\section{Conclusion}
We have introduced Vicinity Vision Transformer (VVT), a general-purpose vision backbone that produces hierarchical feature representations through a pyramid structure.
At its core, we propose Vicinity Attention, which introduces 2D locality into linear self-attention modules. We validate that the locality mechanism improves results and helps generate better class activation.
Then, targeting the computational bottleneck of the linear attention, we propose a novel Vicinity Attention Block to further reduce computation.
In addition, because VVT has linear complexity, it facilitates processing feature maps or input images with higher resolutions.
The proposed VVT has been validated with strong performance on both image classification and semantic segmentation. 
Future work will be aimed at using neural architecture search to further improve upon architectures specifically designed for linear vision transformers.

\bibliographystyle{ieeetr}
\bibliography{main_paper}


\clearpage

\begin{IEEEbiography}[{\includegraphics[width=1in,height=1.25in]{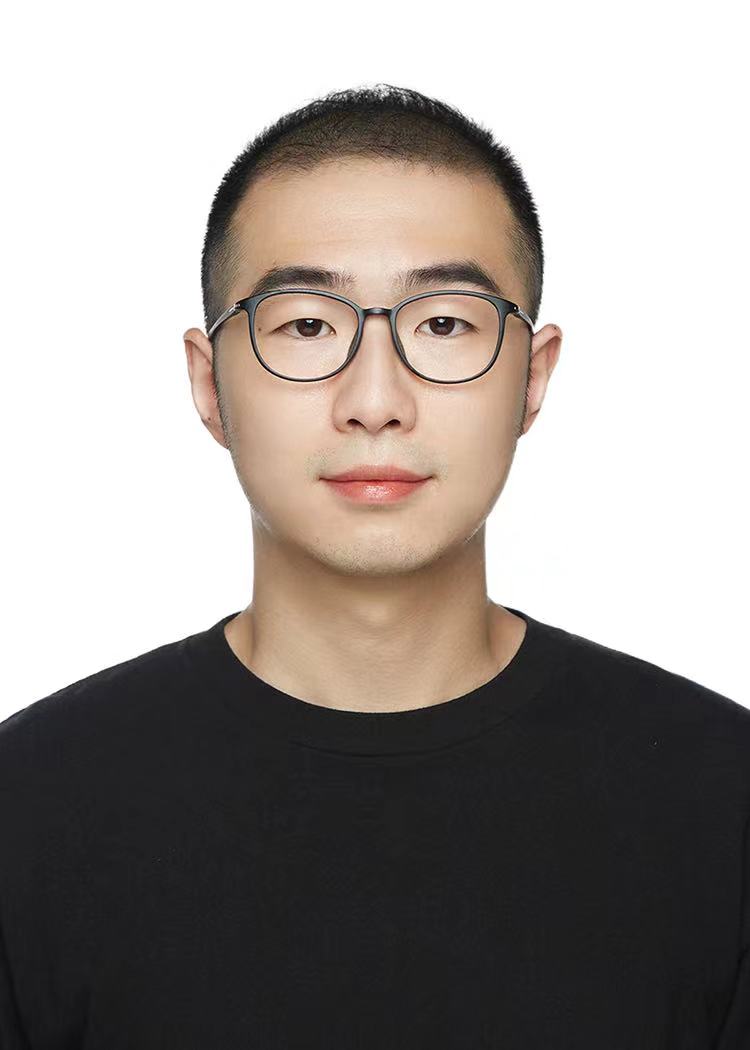}}]{Weixuan Sun} 
Weixuan Sun is a Ph.D student at the Australian National University(ANU), Canberra, Australia.  
He received the double B.E. degrees in Mechatronics engineering and Mechanics engineering from ANU and Beijing Institute of Technology, Beijing, China.
During his Ph.D he joined Sensetime research as a intern to work on Multi-Modal learning. His research interests include weakly-supervised segmentation and efficient transformer.
\end{IEEEbiography}

\begin{IEEEbiography}[{\includegraphics[width=1in,height=1.25in]{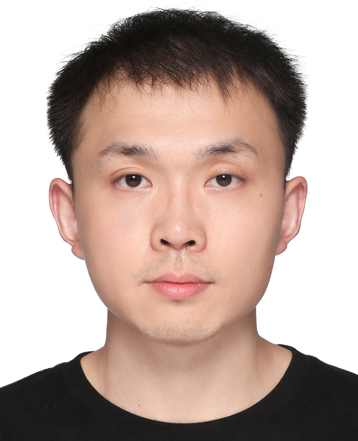}}]{Zhen Qin} 
Zhen Qin is a researcher from Sensetime. He received his B.S. degree in mathematics and applied mathematics from Fudan University in 2017 and his M.S. degree in applied statistics from Tsinghua University in 2020. His current research areas are Efficient Network Design and Transformer.
\end{IEEEbiography}

\begin{IEEEbiography}[{\includegraphics[width=1in,height=1.25in]{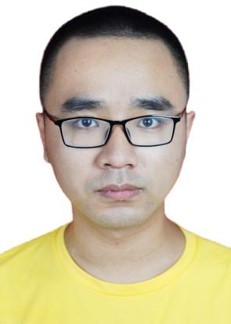}}]{Hui Deng} 
Hui Deng is a Ph.D student at the Northwestern Polytechnical University(NWPU).He received the B.E. degree and M.E degree from Northwestern Polytechnical University Xi'an, China in 2019 and 2022 resprctively . During his Ph.D he joined sensetime research as a research intern to work on MultiModal Machine Learning. His current research interests including gernerative models, non-rigid structure from motion and depth estimation.
\end{IEEEbiography}

\begin{IEEEbiography}[{\includegraphics[width=1in,height=1.25in]{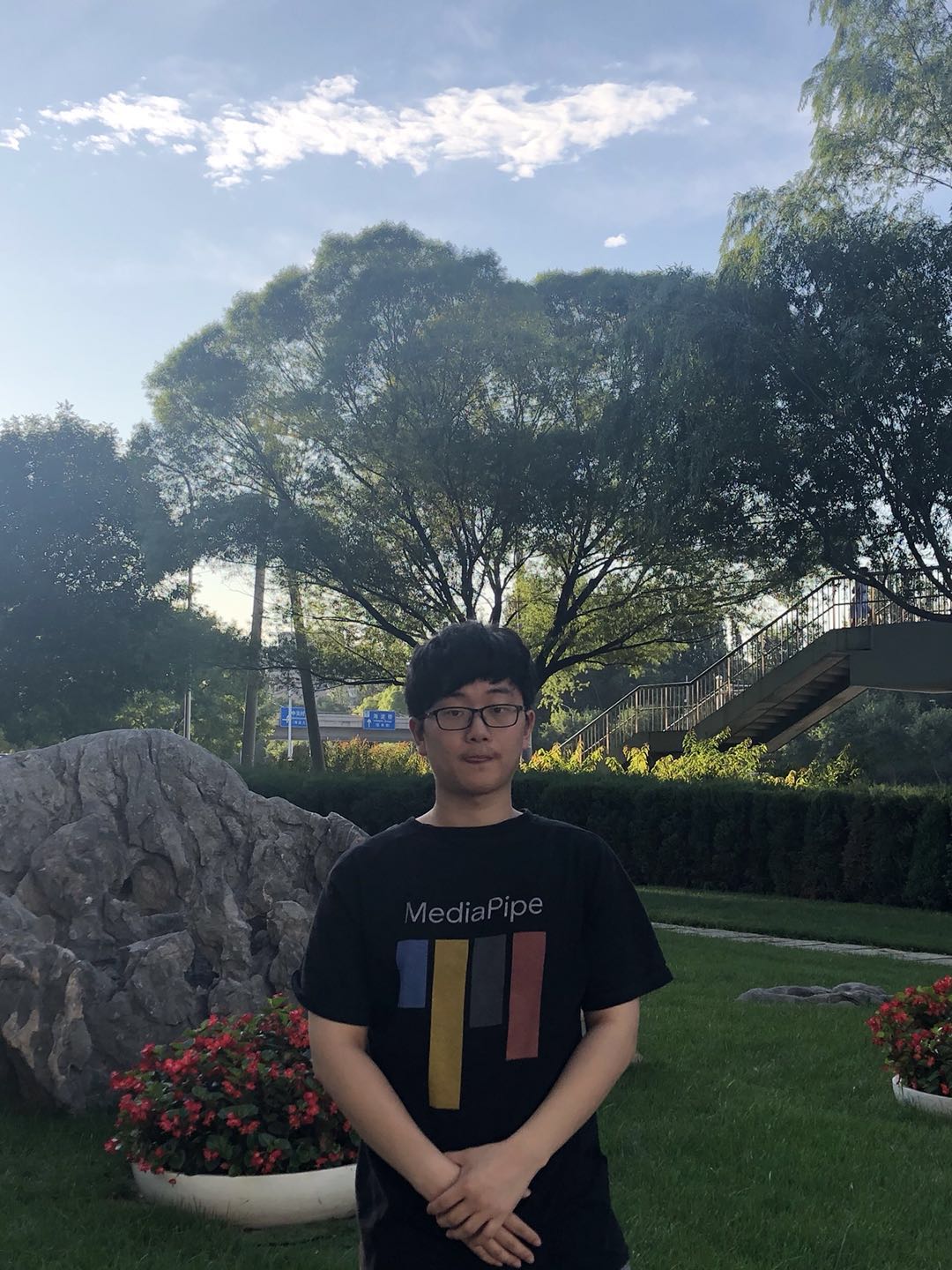}}]{Jianyuan Wang} 
Jianyuan Wang is currently a DPhil student in the Visual Geometry Group, University of Oxford. He achieved his B.E. degree with first-class honors from the Australian National University in 2019. He serves as the reviewer for TPAMI, ICLR, CVPR, ICCV, and so on. His research interests include visual geometry and generative modelling.
\end{IEEEbiography}

\begin{IEEEbiography}[{\includegraphics[width=1in,height=1.25in]{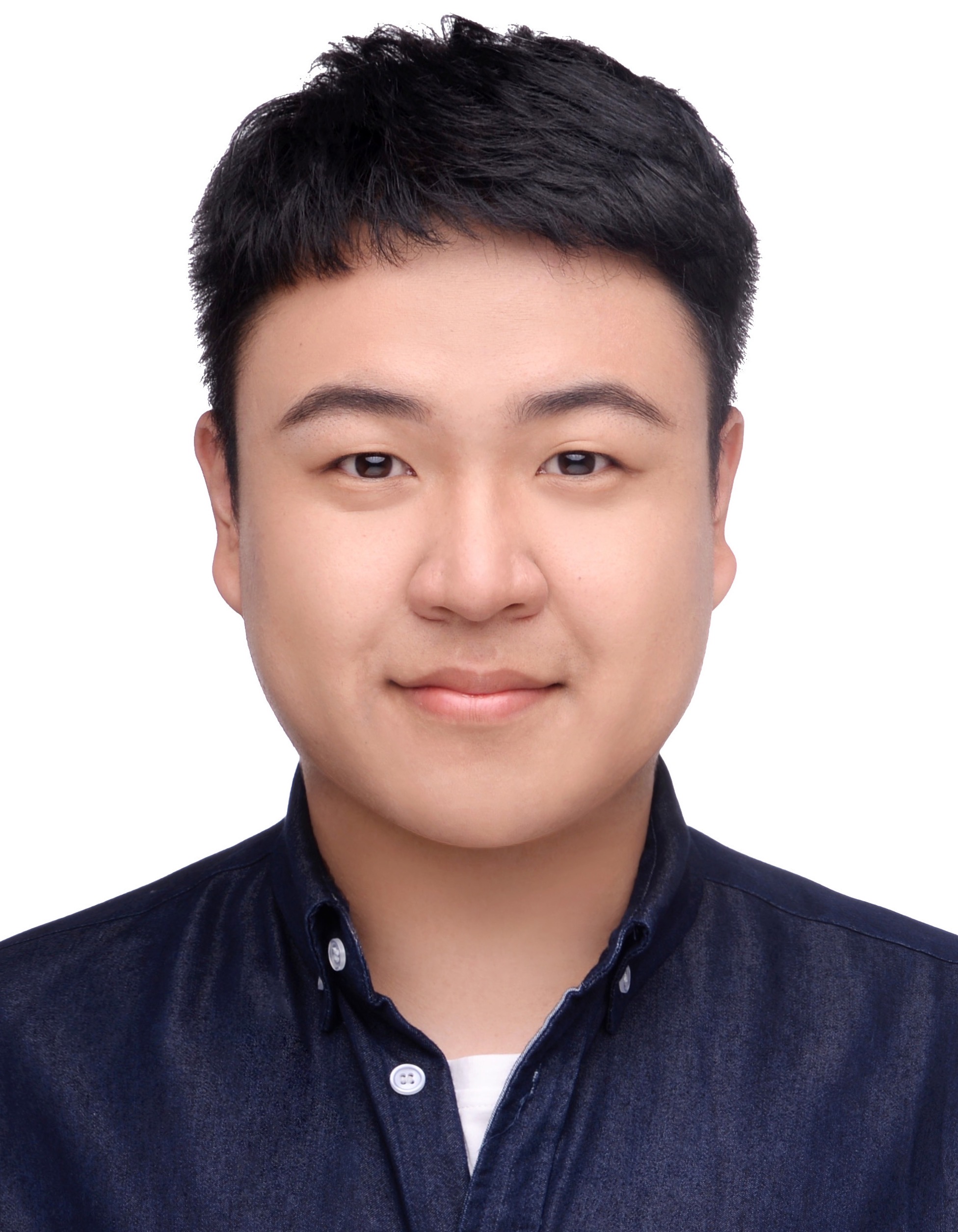}}]{Yi Zhang} 
Yi Zhang is currently a computer vision researcher at SenseTime. He received his M.S. degree in advanced computer science from the University of Birmingham(UOB), Birmingham, UK, in 2019. His research interests include deep learning, computer vision and object detection, especially dense pedestrian detection, cross-modality object detection, and 3D reconstruction.
\end{IEEEbiography}

\begin{IEEEbiography}[{\includegraphics[width=1in,height=1.25in]{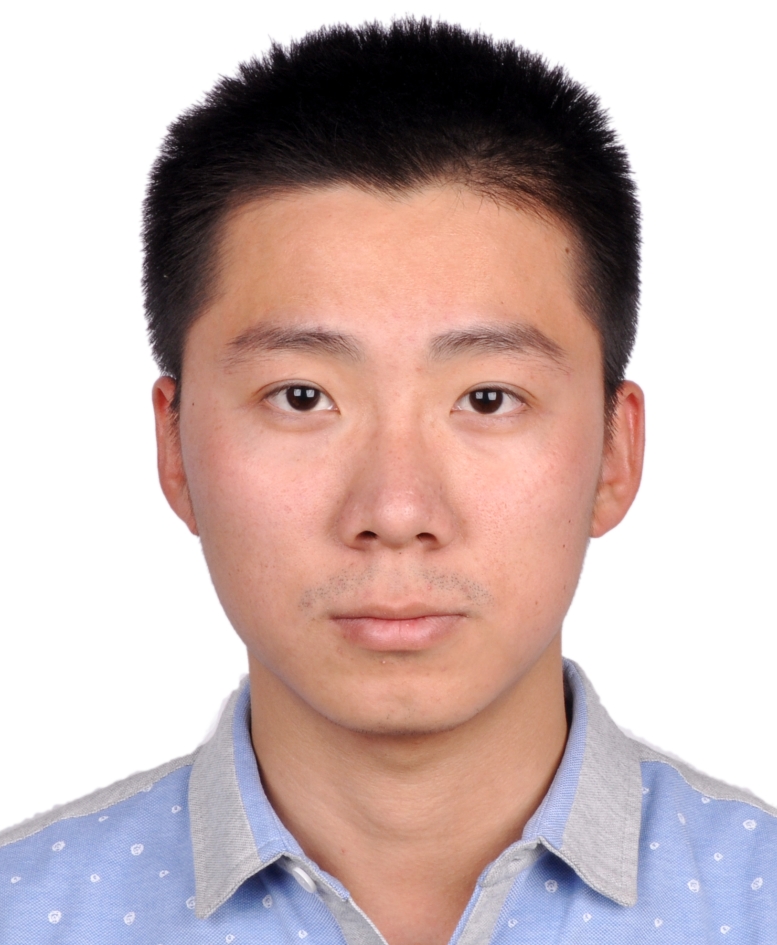}}]{Kaihao Zhang} 
Kaihao Zhang is pursuing the Ph.D. degree with the College of Engineering and Computer Science, The Australian National University, Canberra, ACT, Australia. His research interests focus on computer vision and deep learning. He has more than 30 referred publications in international conferences and journals, including CVPR, ICCV, ECCV, NeurIPS, AAAI, ACMMM, IJCV, TIP, TMM, etc.
\end{IEEEbiography}

\begin{IEEEbiography}[{\includegraphics[width=1in,height=1.25in]{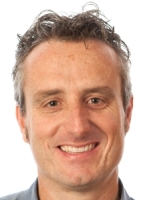}}]{Nick Barnes} 
Nick Barnes is a Professor with
the School of Computing at the Australian National
University. He received his B.Sc. (Hons)
degree and PhD in Robotic Vision
from the University of Melbourne, in 1994 and
1999. 
With NICTA, 2003-16, and CSIRO, he was a Senior Principal Researcher and led the Computer Vision Research Group. 
He has best paper awards and nominations including from
CVPR, Robotics and Systems Science, IROS, and DICTA. 
\end{IEEEbiography}

\begin{IEEEbiography}[{\includegraphics[width=1in,height=1.25in]{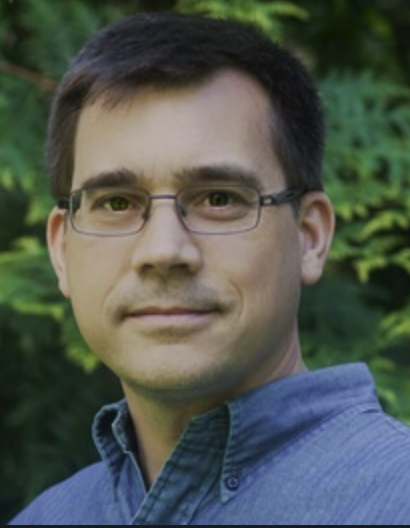}}]{Stan Birchfield} 
Stan Birchfield is a Principal Research Scientist and Senior Research Manager. 
from NVIDIA. He remains an adjunct faculty member at Clemson.  He conducted research at Microsoft, was the principal architect of a commercial product at a startup company in the Bay Area, co-founded a startup with collaborators at Clemson, and consulted for various companies.  He has authored or co-authored more than 100 publications and a textbook on image processing and analysis.  He regularly serves on the program committees and editorial boards of various leading conferences and journals in computer vision and robotics.  He received his Ph.D. in electrical engineering, with a minor in computer science, from Stanford University. 
\end{IEEEbiography}

\begin{IEEEbiography}[{\includegraphics[width=1in,height=1.25in]{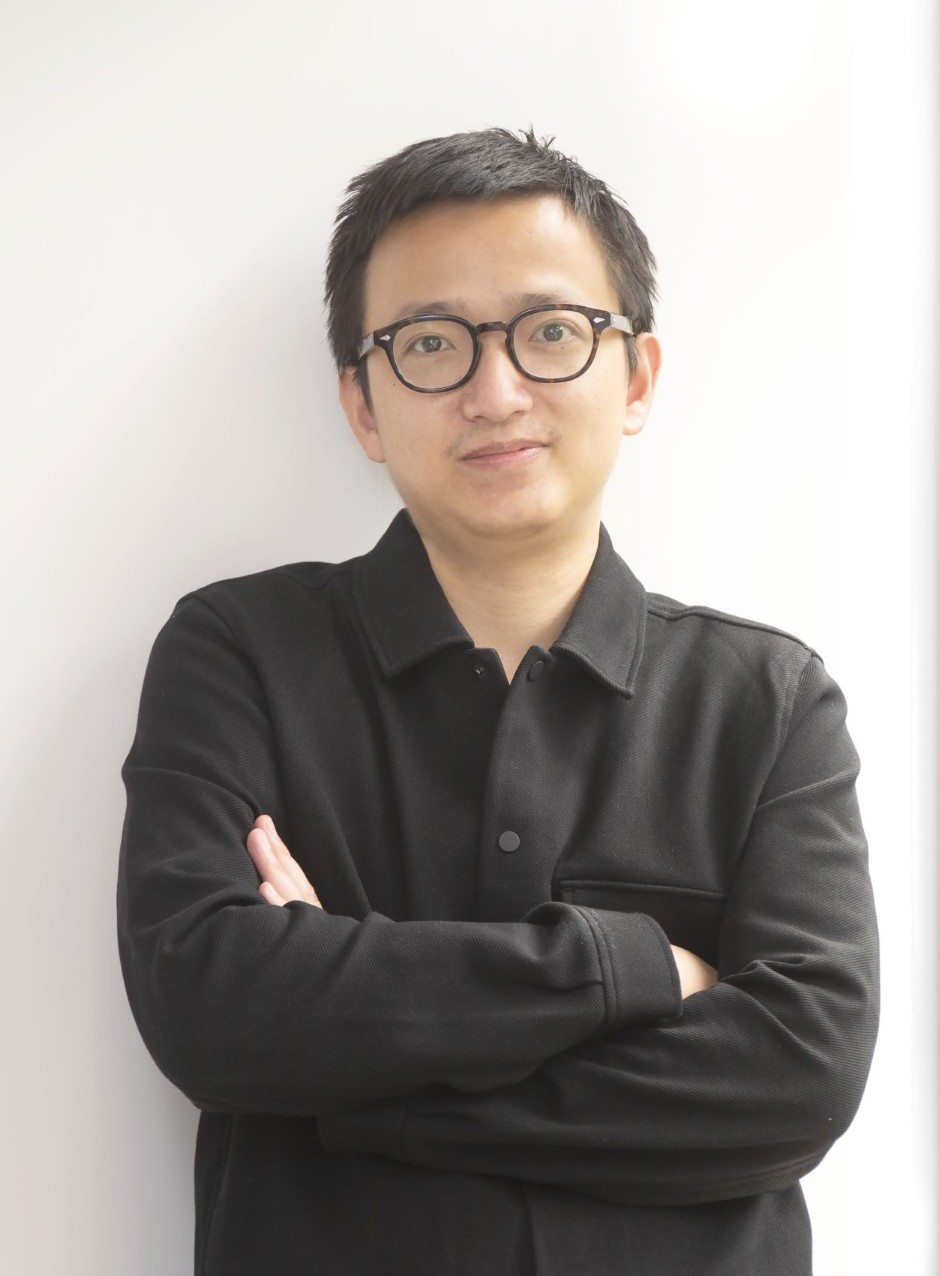}}]{Lingpeng Kong} 
Lingpeng Kong is an assistant professor in the Department of Computer Science at the University of Hong Kong (HKU), and a co-director of the HKU NLP Lab. His work lies at the intersection of natural language processing (NLP) and machine learning (ML), with a focus on representation learning, structured prediction, and generative models. Before joining HKU, he was a research scientist at Google DeepMind. He obtained his Ph.D. from School of Computer Science, Carnegie Mellon University.
\end{IEEEbiography}

\begin{IEEEbiography}[{\includegraphics[width=1in,height=1.25in]{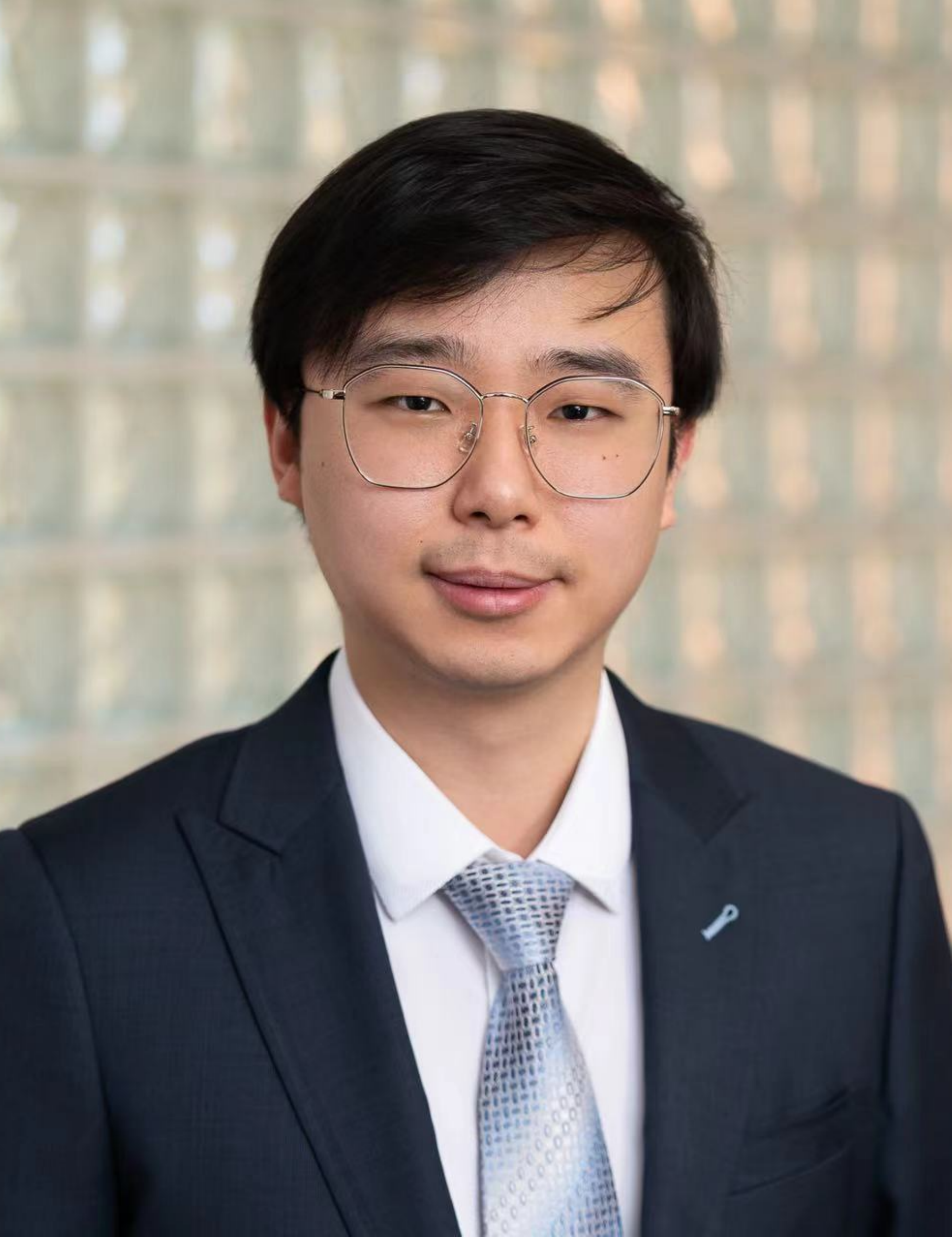}}]{Yiran Zhong} 
Yiran Zhong is currently a principal investigator in Shanghai AI Laboratory. He received a Ph.D. degree of Engineering from The Australian National University, in 2021 and a M.Eng with the first class honor in information and electronics engineering from The Australian National University, in 2014. His research interests include self-supervised learning, visual geometry learning, multimodality leaning, machine learning, and  natural language processing. He won the ICIP Best Student Paper Award in 2014.
\end{IEEEbiography}

\end{document}